# Planning through Stochastic Local Search and Temporal Action Graphs in LPG

**Alfonso Gerevini**                                    GEREVINI@ING.UNIBS.IT
**Alessandro Saetti**                                     SAETTI@ING.UNIBS.IT
**Ivan Serina**                                          SERINA@ING.UNIBS.IT
*Dipartimento di Elettronica per l'Automazione, Università degli Studi di Brescia*
*Via Branze 38, I-25123 Brescia, Italy*

## Abstract

We present some techniques for planning in domains specified with the recent standard language PDDL2.1, supporting "durative actions" and numerical quantities. These techniques are implemented in LPG, a domain-independent planner that took part in the 3rd International Planning Competition (IPC). LPG is an incremental, any time system producing multi-criteria quality plans. The core of the system is based on a stochastic local search method and on a graph-based representation called "Temporal Action Graphs" (TA-graphs). This paper focuses on temporal planning, introducing TA-graphs and proposing some techniques to guide the search in LPG using this representation. The experimental results of the 3rd IPC, as well as further results presented in this paper, show that our techniques can be very effective. Often LPG outperforms all other fully-automated planners of the 3rd IPC in terms of speed to derive a solution, or quality of the solutions that can be produced.

## 1. Introduction

Modeling temporal and numerical information in automated planning is important for representing real-world domains, where actions take time, and consume resources, and the quality of the solutions should take these aspects into account. In the '80s and early '90s some expressive, but inefficient, planning systems handling time were developed (e.g., Vere, 1983; Tsang, 1986; Allen, 1991; Penberthy & Weld, 1994). More recently, a number of alternative interesting approaches to temporal planning has been proposed (e.g., Smith & Weld, 1999; Do & Kambhampati, 2001; Haslum & Geffner, 2001; Dimopoulos & Gerevini, 2002). Some of these planners can compute plans with optimal makespan, but in practice most of them scale up poorly.

Local search is emerging as a powerful method to address fully-automated planning, though in principle this approach does not guarantee generation of optimal plans. In particular, two planners that successfully participated in the recent 3rd International Planning Competition (IPC) are based on local search: FF (Hoffmann & Nebel, 2001) and LPG.

In earlier work on LPG (Gerevini & Serina, 1999, 2002) we proposed a first version of our system using several techniques for local search in the space of *action graphs* (A-graphs), particular subgraphs of the planning graph representation (Blum & Furst, 1997). This version of the planner handled only STRIPS domains, possibly extended with simple costs associated with the actions. In this paper, which is a revised and extended version of a recent work (Gerevini, Serina, Saetti, & Spinoni, 2003), we present some major improvements that were used in the 3rd IPC to handle domains specified in the recent PDDL2.1 language supporting "durative actions" and numerical quantities (Fox & Long, 2003).

The general search scheme of our planner is Walk-plan, a stochastic local search procedure similar to the well-known Walk-sat (Selman, Kautz, & Cohen, 1994). Two of the most important extensions on which we focus in this paper concern the use of *temporal action graphs* (TA-graphs), instead of simple A-graphs, and some new techniques to guide the local search process. In a TA-graph, action nodes are marked with temporal values estimating the earliest time when the corresponding action terminates, while fact nodes are marked with temporal values estimating the earliest time when the corresponding fact becomes true. A set of ordering constraints is maintained during search to handle mutually exclusive actions, and to represent the temporal constraints implicit in the "causal" relations between actions in the current plan.

The new heuristics exploit some reachability information to weigh the elements (TA-graphs) in the search neighborhood that resolve an inconsistency selected from the current TA-graph. The evaluation of these TA-graphs is based on the estimated number of search steps required to reach a solution (a valid plan), its estimated makespan, and its estimated execution cost. LPG is an incremental planner, in the sense that it produces a sequence of valid plans each of which improves the quality of the previous ones. Plan quality is modeled by execution and temporal costs in a flexible way (the user can determine the relative importance of the plan quality criteria).

In the 3rd IPC, our planner demonstrated excellent performance on a large set of test problems in terms of both speed to compute the first solution and quality of the best solution computed by the incremental process. LPG was the fully-automated planner that solved the greatest number of problems, and the one with the highest success ratio between attempted problems and solved problems.

The paper is organized as follows. Section 2 presents the action and plan representation used in the competition version of LPG. Section 3 describes LPG's local search neighborhood, some new heuristics for temporal action graphs, and the techniques for computing the reachability and temporal information used in these heuristics. Moreover, in this section we describe how LPG handles numerical variables and the incremental process to produce good quality plans. Section 4 presents the results of an experimental analysis using the test problems of the 3rd IPC, and illustrating the efficiency of our approach especially for temporal planning. Section 5 gives conclusions, and mentions current and future work. Finally, a collection of appendices describes LPG's algorithm for computing the mutual exclusion relations used during search, and gives details about some of the experimental results presented in Section 4.

## 2. Action and Plan Representation

In this section we introduce our graph-based representations for STRIPS and temporal plans, which can be seen as an elaboration of planning graphs (Blum & Furst, 1997).

### 2.1 Planning Graphs and Actions Graphs

A planning graph is a directed acyclic levelled graph with two kinds of nodes and three kinds of edges. The levels alternate between a fact level, containing fact nodes, and an action level containing action nodes. An action node at a level $t$ represents an action (instantiated operator) that can be planned at time step $t$. A fact node at a level $t$ represents a proposition corresponding to a precondition of one or more actions at time step $t$, or to an effect of one or more actions at time step $t - 1$. The fact nodes of level 1 represent the positive facts of the initial state of the planning problem (every fact that is not mentioned in the initial state is considered false).

In the following, we indicate with $[u]$ the proposition (action) represented by the fact node (action node) $u$. The edges in a planning graph connect action nodes and fact nodes. In particular, an action node $a$ at a level $i$ is connected by: *precondition edges* from the fact nodes of level $i$ representing the preconditions of $[a]$; *add-edges* to the fact nodes of level $i + 1$ representing the positive effects of $[a]$; *delete-edges* to the fact nodes of level $i + 1$ representing the negative effects of $[a]$. Each fact node $f$ at a level $l$ is associated with a *no-op* action node at the same level, which represents a dummy action having $[f]$ as its only precondition and effect.

Two action nodes $a$ and $b$ are marked as mutually exclusive in the graph when one of the actions deletes a precondition of the other (*interference*) or an add-effect of the other (*inconsistent effects*), or when a precondition node of $a$ and a precondition node of $b$ are marked as mutually exclusive (*competing needs*).

Two proposition nodes $p$ and $q$ in a proposition level are marked as exclusive if all ways of making proposition $[p]$ true are exclusive with all ways of making $[q]$ true (each action node $a$ having an add-edge to $p$ is marked as exclusive with each action node $b$ having an add-edge to $q$). When two fact or action nodes are marked as mutually exclusive, we say that there is a *mutex* relation (or simply a mutex) between them.

Given a planning problem $\Pi$, the corresponding planning graph $\mathcal{G}$ can be incrementally constructed level by level starting from level 1 using a polynomial algorithm (Blum & Furst, 1997). The graph construction should reach a propositional level where the goal nodes are present, and there is no mutex relation between them.[1] The *fixed-point level* of the graph is the level from which the nodes and mutex relations at every subsequent level remain the same. Blum and Furst refer to this level as the level where the graph has "leveled off". The mutex relations in the planning graph monotonically decrease with the increase of the levels: a mutex relation holding at a certain level may not hold at the next levels, but it is guaranteed that it holds at all previous levels containing the fact/action nodes involved in the relation. The mutex relations at the fixed-point level of the graph are called *persistent mutex* relations (Fox & Long, 2000), because they hold at every level of the graph.

Without loss of generality, we can assume that the goal nodes of the last level represent the preconditions of the special action $[a_{end}]$, which is the last action in any valid plan, while the fact nodes of the first level represent the effects of the special action $[a_{start}]$, which is the first action in any valid plan.

Our approach to planning uses particular subgraphs of $\mathcal{G}$, called *action graphs*, which represent partial plans.

**Definition 1** An **action graph** (A-graph) for $\mathcal{G}$ is a subgraph $\mathcal{A}$ of $\mathcal{G}$ containing $a_{start}$ and $a_{end}$, and such that, if $a$ is an action node of $\mathcal{G}$ in $\mathcal{A}$, then also the fact nodes of $\mathcal{G}$

---

1. In some cases, when the problem is not solvable, the algorithm identifies that there is no level satisfying these conditions, and hence it detects that the problem is unsolvable.

*corresponding to the preconditions and positive effects of [a] are in $\mathcal{A}$, together with the edges connecting them to a.*

Notice that an action graph can represent an invalid plan for the problem under consideration, since it may contain some *inconsistencies*, i.e., an action with precondition nodes that are not *supported*, or a pair of action nodes involved in a mutex relation. In general, a precondition node $q$ at a level $i$ is supported in an action graph $\mathcal{A}$ of $\mathcal{G}$ if either (i) in $\mathcal{A}$ there is an action node at level $i-1$ representing an action with (positive) effect $[q]$, or (ii) $i = 1$ (i.e., $[q]$ is a proposition of the initial state). An action graph without inconsistencies represents a valid plan and is called a *solution graph*.

**Definition 2** *A* **solution graph** *for $\mathcal{G}$ is an action graph $\mathcal{A}_s$ of $\mathcal{G}$ such that all precondition nodes of the actions in $\mathcal{A}_s$ are supported, and there is no mutex relation between action nodes of $\mathcal{A}_s$.*

For large planning problems the construction of the planning graph can be computationally very expensive, especially because of the high number of mutex relations. For this reason our planner considers only pairs of actions that are persistently mutex, derived using a dedicated algorithm given in Appendix A. An experimental comparison with IPP's implementation of the planning graph construction (Koehler, Nebel, Hoffmann, & Dimopoulos, 1997) showed that in practice our method for deriving mutex relations is considerably more efficient than the "traditional" method for deriving the mutex relations in the fixed-point level of the graph. Moreover, for the problems that we tested, our method derived all the persistent mutex relations found by the traditional method.

The definition of action graphs and the notion of supported facts can be made stronger by observing that the effects of an action node can be automatically propagated to the next levels of the graph through the corresponding no-ops, until there is an interfering action *blocking* the propagation (if any), or the last level of the graph has been reached. The use of the no-op propagation, that we presented in previous work (Gerevini & Serina, 2002), leads to a smaller search space and can be incorporated into the definition of action graph.

**Definition 3** *An* **action graph with propagation** *is an action graph $\mathcal{A}$ such that if a is an action node of $\mathcal{A}$ at a level $l$, then, for any positive effect $[e]$ of $[a]$ and any level $l' > l$ of $\mathcal{A}$, the no-op of $e$ at level $l'$ is in $\mathcal{A}$, unless there is another action node at a level $l''$ ($l \leq l'' < l'$) which is mutex with the no-op.*

Since in the rest of this paper we consider only action graphs with propagation, we will abbreviate their name simply to action graphs (leaving implicit that they include the no-op propagation).

In most of the existing planners based on planning graphs, when the search for a solution graph fails, $\mathcal{G}$ is iteratively expanded by adding an extra level and performing a new search using the resulting graph. In systematic planners like GRAPHPLAN (Blum & Furst, 1997), STAN (Fox & Long, 1998b) and IPP (Koehler et al., 1997) the search fails when there exists no solution graph, while in planners that use local search like BLACKBOX (Kautz & Selman, 1999) or GPG (Gerevini & Serina, 1999) the search fails when a certain search limit is exceeded. As we will show, in LPG there is no need to explicitly treat this kind of search failure, since the size of the graph is incrementally increased during search (i.e., the graph extension can be part of a local search step).

## 2.2 Linear and Temporal Action Graphs

The first version of LPG (Gerevini & Serina, 2002) was based on action graphs where each level may contain an arbitrary number of action nodes, as in the usual definition of planning graph. The version of the system that participated in the 3rd IPC uses a restricted class of action graphs, called *linear action graphs*, combined with some additional data structures supporting a more expressive action and plan representation. In particular, the new system can handle actions having temporal durations and preconditions/effects involving numerical quantities, as specified in PDDL2.1 (Fox & Long, 2003). In this paper we focus mainly on planning for temporal domains, where LPG showed particularly good performance with respect to the other (fully-automated) participants of the 3rd IPC.

In order to keep the presentation simple, we describe our techniques considering mainly preconditions of type "over all" (i.e., preconditions that must hold during the whole action execution) and effects of type "at end" (i.e., effects that hold at the end of the action execution).[2] In Section 3.4 we discuss how we handle the other types of preconditions and effects in the test domains of the 3rd IPC.

**Definition 4** *A **linear action graph** (LA-graph) of $\mathcal{G}$ is an A-graph of $\mathcal{G}$ in which each level of actions contains at most one action node representing a domain action and any number of no-ops.*

It is important to note that having only one action in each level of an LA-graph does not prevent the generation of parallel (partially ordered) plans. In fact, from any LA-graph we can easily extract a partially ordered plan where the ordering constraints are (1) those between mutex actions and (2) those implicit in the causal structure of the represented plan. Regarding the first constraints, if $a$ and $b$ are mutex and the level of $a$ precedes the level of $b$, then $[a]$ is ordered before $[b]$; regarding the second constraints, if $a$ has an effect node that is used (possibly through the no-ops) to support a precondition node of $b$, then $[a]$ is ordered before $[b]$. These causal relations between actions producing an effect and actions consuming it are similar to the causal links in partial-order planning (e.g., McAllester & Rosenblitt, 1991; Penberthy & Weld, 1992; Nguyen & Kambhampati, 2001). LPG keeps track of these relationships during search and uses them to derive some heuristic information useful for guiding the search (more details on this in the next section), as well as to extract parallel plans from the solution graph.

For temporal domains where actions have durations and plan quality mainly depends on the makespan, rather than on the number of actions or graph levels, the distinction between one action or more actions per level is scarcely relevant. The order of the graph levels should not imply any ordering of the actions (e.g., an action at a certain level could terminate before the end of an action at the next level).

Since in LA-graphs there is at most one action node for each level, and every inconsistency is an unsupported precondition, the use of this representation has some advantages over general A-graphs:

- LA-graphs can be represented by simpler data structures, which allow one to manage the no-op propagation, the inconsistency identification and selection, and the numerical effect propagation more efficiently.

---

2. LPG supports all types of preconditions and effects that can be expressed in PDDL2.1 (levels 1–3).

- LA-graphs better support the computation of the heuristic and reachability information used by the local search algorithm presented in the next section. As we will see, for these techniques it is important to derive a consistent and possibly complete description of the state where any action in the current plan is applied. In an LA-graph, we can efficiently derive these state descriptions by using the levels of the graph as a total order of the actions in the current plan.

- In numerical domains, if mutex actions can belong to the same level of the current action graph, it could be impossible to determine whether a numerical precondition of an action at a following level is satisfied.[3] In an LA-graph (persistent) mutex actions belong to different levels and are ordered, making this easy to determine.

Also note that the fact of having only one action per level allows us to define a larger search neighborhood. In general, a disadvantage of LA-graphs with respect to A-graphs is the size of the representation, since the number of levels in an LA-graph can be significantly larger than the number of levels in the corresponding A-graph. However, in all planning problems that we tested, the size of LA-graphs was never a problem for our planner.[4]

For PDDL2.1 domains involving durative actions, our planner represents temporal information by an assignment of real values to the action and fact nodes of the LA-graph, and by a set $\Omega$ of *ordering constraints* between action nodes. The value associated with a fact node $f$ ($Time(f)$) represents the earliest time when $[f]$ becomes true, given the actions in the represented plan and the constraints in $\Omega$; the value associated with an action node $a$ ($Time(a)$) represents the earliest time when the execution of $[a]$ can terminate. These temporal values are derived from the duration of the actions in the LA-graph and the ordering constraints between them that are stated in $\Omega$.

**Definition 5** *A temporal action graph* (TA-graph) *of* $\mathcal{G}$ *is a triple* $\langle \mathcal{A}, \mathcal{T}, \Omega \rangle$ *where*

- $\mathcal{A}$ *is a linear action graph;*

- $\mathcal{T}$ *is an assignment of real values to the fact and action nodes of* $\mathcal{A}$;

- $\Omega$ *is a set of ordering constraints between action nodes of* $\mathcal{A}$.

The ordering constraints in a TA-graph are of two types: constraints between actions that are implicitly ordered by the causal structure of the plan ($\prec_C$-*constraints*), and constraints that are imposed by the planner to deal with mutually exclusive actions ($\prec_E$-*constraints*). $a \prec_C b$ belongs to $\Omega$ if and only if $a$ is used to achieve a precondition node of $b$ in $\mathcal{A}$, while $a \prec_E b$ (or $b \prec_E a$) belongs to $\Omega$ only if $a$ and $b$ are mutually exclusive in $\mathcal{A}$ ($a \prec_E b$, if the level of $a$ precedes the level of $b$, $b \prec_E a$ otherwise). In Section 3.4 we will discuss how ordering constraints are introduced by LPG during the search. Given our assumption on the types of action preconditions and effects in temporal domains, an

---

3. For instance, suppose we have an A-graph with two mutex actions at a level such that one action sets the value of the numerical variable $x$ to 10, while the other sets it to 20. Unless we order these actions, it is impossible to determine whether $x > 15$ holds when the action at the next level is applied.

4. LPG's implementation of LA-graphs uses an extended version of Hoffmann's "connectivity graph", a compact representation of the action and fact nodes in a planning graph (Hoffmann & Nebel, 2001). The extensions are needed to represent persistent mutex relations, durative actions and numerical preconditions/effects.

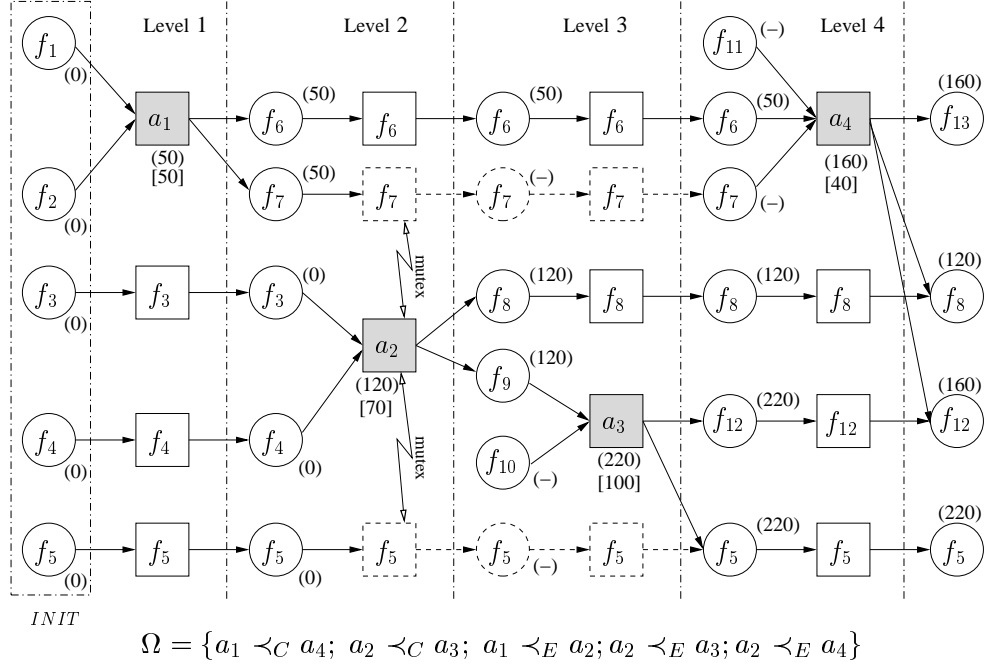

$$\Omega = \{a_1 \prec_C a_4; \ a_2 \prec_C a_3; \ a_1 \prec_E a_2; a_2 \prec_E a_3; a_2 \prec_E a_4\}$$

Figure 1: An example of TA-graph. Dashed edges form chains of no-ops that are blocked by mutex actions. Round brackets contain temporal values assigned by $\mathcal{T}$ to the fact nodes (circles) and the action nodes (squares). The numbers in square brackets represent the durations of the actions. "(–)" indicates that the corresponding fact node is not supported.

ordering constraint $a \prec b$ (where "$\prec$" stands for $\prec_C$ or $\prec_E$) states that the end of $[a]$ is before the start of $[b]$. The temporal value assigned by $\mathcal{T}$ to a node $x$, denoted by $Time(x)$, is derived as follows. If a fact node $f$ of the action graph is unsupported, then $Time(f)$ is undefined, otherwise it is the minimum value over the temporal values assigned to the actions supporting it. If the temporal value of every precondition node of an action node $a$ is undefined, and there is no action node with a temporal value that must precede $a$ according to $\Omega$, then $Time(a)$ is set to the duration of $a$; otherwise $Time(a)$ is the sum of the duration of $a$ and the maximum value over the temporal values of its precondition nodes and the temporal values of the action nodes that must precede $a$.

Figure 1 gives an example of a TA-graph containing four action nodes ($a_{1...4}$) and several fact nodes representing thirteen facts. Since $a_1$ supports a precondition node of $a_4$, $a_1 \prec_C a_4$ belongs to $\Omega$ (similarly for $a_2 \prec_C a_3$). $a_1 \prec_E a_2$ also belongs to $\Omega$ because $a_1$ and $a_2$ are persistently mutex (similarly for $a_2 \prec_E a_3$ and $a_2 \prec_E a_4$). The temporal value assigned to the facts $f_{1...5}$ at the first level is zero, because they belong to the initial state. $a_1$ has all its preconditions supported at time zero, and hence $Time(a_1)$ is the duration of $a_1$. Since $a_1 \prec a_2 \in \Omega$, $Time(a_2)$ is given by the sum of the duration of $a_2$ and the maximum value over the temporal values of its precondition nodes (zero) and $Time(a_1)$. Since $f_9$ at level 3 is supported only by $a_2$, and this is the only supported precondition node of $a_3$, $Time(a_3)$ is the sum of $Time(a_2)$ = $Time(f_9)$ and the duration of $a_3$. Since $a_2$ must precede $a_4$ (but there is no ordering constraint between $a_3$ and $a_4$), $Time(a_4)$ is the maximum value over $Time(a_2)$ and the temporal values of its supported precondition nodes ($f_6$), plus the duration of $a_4$. Finally, note that $f_{12}$ at the last level is supported both by $a_4$ and $a_3$. Since $Time(a_3) > Time(a_4)$, we have that $Time(f_{12})$ at this level is equal to $Time(a_4)$.

**Definition 6** *A temporal solution graph for $\mathcal{G}$ is a TA-graph $\langle \mathcal{A}, \mathcal{T}, \Omega \rangle$ such that $\mathcal{A}$ is a solution LA-graph of $\mathcal{G}$, $\mathcal{T}$ is consistent with $\Omega$ and the duration of the actions in $\mathcal{A}$, $\Omega$ is consistent, and for each pair $\langle a, b \rangle$ of mutex actions in $\mathcal{A}$, either $\Omega \models a \prec b$ or $\Omega \models b \prec a$.*

While obviously the levels in a TA-graph do not correspond to real time values, they represent a topological order for the $\prec_C$-constraints in the TA-graph (i.e., the actions of the TA-graph that are ordered according to their relative levels form a linear plan satisfying all $\prec_C$-constraints). This topological sort can be a valid total order for the $\prec_E$-constraints of the TA-graph as well, provided that these constraints are appropriately stated during search, i.e., that if $a$ and $b$ are exclusive, the planner appropriately imposes either $a \prec_E b$ or $b \prec_E a$. LPG chooses $a \prec_E b$ if the level of $a$ precedes the level of $b$, $b \prec_E a$ otherwise. Under this assumption on the "direction" in which $\prec_E$-constraints are imposed, it is easy to see that the levels of a TA-graph correspond to a topological order of the actions in the represented plan satisfying every ordering constraint in $\Omega$.

For planning domains that require minimizing the plan makespan (like the "Time", "SimpleTime", "Complex", and some of the "Numeric" and "HardNumeric" domain sets of the 3rd IPC) each element of LPG's search space is a TA-graph. For domains where time is irrelevant (like the "Strips" and "Numeric" domain sets of the 3rd IPC) the search space is formed by LA-graphs.[5]

## 2.3 Action Durations and Costs

In this section we comment on the representation of action durations and action costs in LPG. In accordance with PDDL2.1, our planner handles both static durations and dynamic durations, i.e., durations depending on the state in which the action is applied. Static durations are either explicitly given as numbers specified in the field ":duration" of the operator description, or they are implicitly specified by an expression involving some static quantities specified in the initial state of the planning problem. An example of implicit static duration is the duration of the Drive actions in the Depots-Time domain of the 3rd IPC: the Drive operator defines the duration as the distance between the source and the destination of travel (two operator parameters instantiated by values specified in the initial state), divided by the speed of the vehicle that is driven (another operator parameter).

Typically, dynamic durations depend on some numeric quantities that may vary from one state to another state reached by the actions in the plan. An example is the energy of a rover in the domain Rovers-Time of the 3rd IPC, where the duration specified in the recharge operator is

    (/ (- 80 (energy ?x)) (recharge-rate ?x))).

---

5. An experimental analysis showed that in STRIPS domains the techniques for LA-graphs are more powerful than the techniques for A-graphs that we proposed in previous work (Gerevini & Serina, 2002).

This expression depends on the current value of **energy** for the rover **?x** and on its static recharging rate (**recharge-rate**) specified in the initial state. Our planner handles the dynamic duration of an action by computing and maintaining during search an estimate of the value of the numerical quantities in the state where the action is applied. In Section 3.5 we will briefly describe how the version of LPG that took part in the 3rd IPC handles numerical state variables, and numerical preconditions and effects involving them. However, in this paper we will not describe their treatment in detail, and in the next section we will assume that action durations are static.

Each action of a plan can be associated with a cost that may affect the plan quality. Like action durations, in general these costs could be either static or dynamic, though the current version of LPG handles only static ones. LPG precomputes the action costs using the plan metric specified in the problem description using the PDDL2.1 field ":metric".[6] For instance, the plan metric used for a problem in the **ZenoTravel-Numeric** domain of the 3rd IPC is

(:metric minimize (+ (* 4 (total-time)) (* 5 (total-fuel-used)))),

i.e., it is the sum of four times the plan makespan and five times the total amount of the fuel used by the actions in the plan. The cost of an action $a$ is derived by evaluating how the value of the plan metric expression is changed by the effects of $a$. LPG computes an initial value $m_0$ for the expression by using the values specified in the initial state as values of the involved numerical variables. From $m_0$ LPG derives a new value $m_1$ by applying the effects of $a$ that increase/decrease the value of one or more variables in the expression. The cost of $a$ is defined as $m_1 - m_0$. If this difference is zero, in order to prefer plans containing a lower number of actions, the cost of $a$ is set to a small positive quantity. Notice also that in these evaluations of the metric expression the temporal value **total-time** is not considered (it is set to zero, if present), because the temporal aspect of the plan quality is already taken into account by the durations of the actions. In the previous example, the metric subexpression used to derive the action costs is (* 5 (total-fuel-used)). Thus, for instance, the cost of the **ZenoTravel** action (fly plane1 city0 city1) in the problem **pfile1** of the 3rd IPC is 13560 because the effects of this action increase **total-fuel-used** by the following quantity

(* (distance city0 city1) (slow-burn plane1))) = 678 * 4 = 2712,

which increases the metric value of the plan by 5 * 2712 = 13560.

## 3. Local Search in the Space of Temporal Action Graphs

In this section we present some search techniques used in LPG. We start with a description of the general local search scheme in the space of action graphs. Then we concentrate on temporal action graphs giving a detailed description of LPG's heuristics and of its methods for computing and using reachability information, for maintaining the TA-graph representation during search, and for deriving good quality plans incrementally. In order to simplify the notation, instead of using $a$ and $[a]$ to indicate an action node and the action repre-sented by this node respectively, we will use $a$ to indicate both of them (the appropriate interpretation will be clear from the context).

---

6. As in the domains of the competitions, we assume that the plan metric expression is linear. For simple STRIPS domains, where there is no metric expression to minimize, the cost of each action is set to one, and LPG minimizes the number of actions in the plan.

## 3.1 Basic Search Procedure: Walkplan

Given a planning graph $\mathcal{G}$, the local search process of LPG starts from an initial A-graph of $\mathcal{G}$ (i.e., a partial plan), and transforms it into a solution graph (i.e., a valid plan) through the iterative application of graph modifications improving the current partial plan. The two basic modifications consist of an extension of the A-graph to include a new action node, or a reduction of the A-graph to remove an action node (and the relevant edges).[7] At any step of the search process, which produces a new A-graph, the set of actions that can be added or removed is determined by the inconsistencies that are present in the current A-graph.

The general scheme for searching for a solution graph (a final state of the search) consists of two main steps. The first step is an initialization of the search in which we construct an initial A-graph. The second step is a local search process in the space of all A-graphs, starting from the initial A-graph. We can generate an initial A-graph in several ways. Four possibilities that can be performed in polynomial time, and that we have implemented are: an empty A-graph (i.e., containing only the no-ops of the facts in the initial state, and the special action nodes $a_{start}$ and $a_{end}$); a randomly generated A-graph; an A-graph where all precondition facts are supported, but in which there may be some violated mutex relations; and an A-graph obtained from an existing plan given as input to the process. The last option is particularly useful in the plan optimization phase, as well as for solving plan adaptation problems (Gerevini & Serina, 2000). In the current version of LPG, the default initialization strategy is the empty action graph with the fixed-point level as the last level of the graph. Further details on the initialization step can be found in earlier papers on planning through local search and action graphs (Gerevini & Serina, 1999, 2000).

Once we have computed an initial A-graph, each basic search step selects an inconsistency in the current A-graph. If this is an unsupported fact node, then in order to resolve (eliminate) it, we can either add an action node that supports it, or we can remove an action node that is connected to that fact node by a precondition edge. If the chosen inconsistency is a mutex relation, then we can remove one of the action nodes of the mutex relation. Note that the elimination of an action node can remove several inconsistencies (e.g., all those corresponding to the unsupported preconditions of the action removed). On the other hand, obviously the addition of an action node can introduce several new inconsistencies. The strategy for selecting the next inconsistency to handle may have a significant impact on the overall performance (this has been extensively studied in the context of causal-link partial-order planning, e.g., Pollack, Joslin, & Paolucci, 1997; Gerevini & Schubert, 1996). Our planner includes several strategies that we are currently testing. The default strategy that we used in the 3rd IPC and in all experiments presented in Section 4, prefers inconsistencies appearing at the earliest level of the graph.

Given an action graph $\mathcal{A}$ and an inconsistency $\sigma$ in $\mathcal{A}$, the *neighborhood* $N(\sigma, \mathcal{A})$ of $\sigma$ in $\mathcal{A}$ is the set of A-graphs obtained from $\mathcal{A}$ by applying a graph modification that resolves

---

7. Another possible modification that is analyzed by Gerevini and Serina (2002), but that will not be considered in this paper, is action ordering, i.e., moving forward or backward one of two exclusive action nodes.

Walkplan($\Pi$, $max\_steps$, $max\_restarts$, $p$)

   *Input*: A planning problem $\Pi$, the maximum number of search steps $max\_steps$,
        the maximum number of search restarts $max\_restarts$, a noise factor $p$ ($0 \leq p \leq 1$).
   *Output*: A solution graph representing a plan solving $\Pi$ or `fail`.

1. **for** $i \leftarrow 1$ *to* $max\_restarts$ **do**
2.     $\mathcal{A} \leftarrow$ an initial A-graph derived from the planning graph of $\Pi$;
3.     **for** $j \leftarrow 1$ **to** $max\_steps$ **do**
4.         **if** $\mathcal{A}$ is a solution graph **then**
5.            **return** $\mathcal{A}$
6.         $\sigma \leftarrow$ an inconsistency in $\mathcal{A}$;
7.         $N(\sigma, \mathcal{A}) \leftarrow$ neighborhood of $\mathcal{A}$ for $\sigma$;
8.         **if** $\exists \mathcal{A}' \in N(\sigma, \mathcal{A})$ such that the quality of $\mathcal{A}'$ is not worse than the quality of $\mathcal{A}$
9.            **then** $\mathcal{A} \leftarrow \mathcal{A}'$ (if there is more than one $\mathcal{A}'$-graph, choose randomly one)
10.         **else if** `random` $< p$ **then**
11.            $\mathcal{A} \leftarrow$ an element of $N(\sigma, \mathcal{A})$ randomly chosen
12.            **else** $\mathcal{A} \leftarrow$ best element in $N(\sigma, \mathcal{A})$;
13. **return** `fail`.

Figure 2: General scheme of Walkplan with restarts. `random` is a randomly chosen value between 0 and 1. The quality of an action graph in the neighborhood is measured using an evaluation function estimating the cost of the graph modification used to generate it from the current action graph.

$\sigma$. At each step of the local search scheme, the elements of the neighborhood are evaluated according to a function estimating their quality, and an element with the best quality is then chosen as the next possible A-graph (search state). The quality of an A-graph depends on a number of factors, such as the number of inconsistencies and the estimated number of search steps required to resolve them, the overall cost of the actions in the represented plan and its makespan.[8]

   Gerevini and Serina (1999) proposed three general strategies for guiding the local search: Walkplan, Tabuplan and T-Walkplan. In this paper we focus on Walkplan, which is the strategy used by LPG in the 3rd IPC, as well as in the experimental tests presented in Section 4. Walkplan is similar to Walksat, a stochastic local search method for solving propositional satisfiability problems (Selman et al., 1994; Kautz & Selman, 1996). In Walkplan the best element in the neighborhood is the A-graph which has the *lowest decrease of quality* with respect to the current A-graph, i.e., it does not consider possible improvements. Like Walksat, our strategy uses a *noise parameter* $p$. Given an A-graph $\mathcal{A}$ and an inconsistency $\sigma$, if there is a modification for $\sigma$ that does not decrease the quality of $\mathcal{A}$, then this modification is performed, and the resulting A-graph is chosen as the next A-graph; otherwise, with probability $p$ one of the graphs in $N(\sigma, \mathcal{A})$ is chosen randomly, and with probability $1 - p$ the next A-graph is chosen according to the minimum value of the evaluation function. If a solution graph is not reached after a certain number of search steps ($max\_steps$), the current A-graph and $max\_steps$ are reinitialized, and the search is repeated up to a user-defined maximum number of times ($max\_restarts$). Figure 2 gives a formal description of Walkplan with restarts.

---

8. For simple STRIPS domains the execution cost of the plan is measured in terms of the number of actions (i.e., each action has cost 1), while plan makespan is ignored. Alternatively it can be modeled as the number of parallel time steps (Gerevini & Serina, 2002).

Gerevini and Serina (2002) proposed some heuristic functions for evaluating the search neighborhood of A-graphs with action costs. In the next section we present additional, more powerful heuristic functions for LA-graphs and TA-graphs. These techniques are implemented in the latest version of our planner and were used in the 3rd IPC.

## 3.2 Neighborhood and Heuristics for Temporal Action Graphs

The search neighborhood for an inconsistency $\sigma$ in an LA-graph $\mathcal{A}$ is the set of LA-graphs that can be derived from $\mathcal{A}$ by adding an action node supporting $\sigma$, or removing the action with precondition $\sigma$ (in linear graphs the only type of inconsistencies are unsupported preconditions). An action $a$ supporting $\sigma$ can be added to $\mathcal{A}$ at any level $l$ preceding the level of $\sigma$, and such that the desired effect of $a$ is not blocked before or at the level of $\sigma$ (assuming that the underlying planning graph contains $a$ at level $l$). The neighborhood for $\sigma$ contains a linear action graph for each of these possibilities.

Since at any level of an LA-graph there can be at most one action node (plus any number of no-ops), when we remove an action node from $\mathcal{A}$, the corresponding action level becomes "empty" (i.e., it contains only no-ops).[9] If the LA-graph contains adjacent empty levels, and in order to resolve the selected inconsistency a certain action node can be added at any of these levels, then the corresponding neighborhood contains only one of the resulting graphs.

When we add an action node to a level $l$ that is not empty, the LA-graph is extended by one level, all action nodes from $l$ are shifted forward by one level, and the new action is inserted at level $l$ (Figure 8 in Section 3.4 gives an example). Moreover, when we remove an action node $a$ from the current LA-graph, we can also remove each action node $b$ supporting only the preconditions of $a$. Similarly, we can remove the actions supporting only the preconditions of $b$, and so on. While this induced pruning is not necessary, an experimental analysis showed that it tends to produce better quality plans more quickly.

The elements of the neighborhood are evaluated according to an *action evaluation function* $E$ estimating the cost of adding $(E(a)^i)$ or removing an action node $a$ $(E(a)^r)$. In general, $E$ consists of three weighed terms evaluating three aspects of the quality of the current plan that are affected by the addition/removal of $a$:

$$E(a) = \begin{cases} E(a)^i = \alpha \cdot Execution\_cost(a)^i + \beta \cdot Temporal\_cost(a)^i + \gamma \cdot Search\_cost(a)^i \\ \\ E(a)^r = \alpha \cdot Execution\_cost(a)^r + \beta \cdot Temporal\_cost(a)^r + \gamma \cdot Search\_cost(a)^r \end{cases}$$

The first term of $E$ estimates the increase of the plan execution cost ($Execution\_cost$), the second estimates the end time of $a$ ($Temporal\_cost$), and the third estimates the increase of the number of search steps needed to reach a solution graph ($Search\_cost$). The coefficients of these terms are used to normalize them, and to weigh their relative importance (more on this in Section 3.6).

---

9. Note that the empty levels are ignored during the extraction of the plan from the (temporal) solution graph.

In the computation of the terms of $E$ there is an important tradeoff to consider. On one hand, an accurate evaluation of them could lead to valid plans of good quality within few search steps. On the other hand, the computation of $E$ should be fast "enough", because the neighborhood could contain many elements, and an accurate evaluation of its elements could slow down the search excessively. In the design of our heuristics for evaluating the terms of $E$ we took this tradeoff into account trying to find an appropriate balance between informativeness and efficiency of computation.

The evaluation of the terms of $E$ is based on computing particular relaxed plans for achieving certain action preconditions in the context of the current TA-graph. In the next subsections, first we describe how these relaxed plans are derived, and then we give a detailed description of how the terms of $E$ are defined using relaxed plans.

### 3.2.1 RELAXED PLANS FOR ACTION PRECONDITIONS

Suppose we are evaluating the addition of $a$ at a level $l$ of the current linear action graph $\mathcal{A}$. The three terms of $E$ are heuristically estimated by computing a relaxed plan $\pi_r$ containing a minimal set of actions for achieving (1) the unsupported preconditions of $a$ and (2) the set $\Sigma$ of preconditions of the other actions in the LA-graph that would become unsupported by adding $a$ (because it would block the no-op propagation currently used to support such preconditions). This plan is relaxed because during its construction we do not consider the possible interference between actions resulting from delete-effects.

$\pi_r$ is computed in two stages. First we deal with the preconditions of type (1) and then with the preconditions of type (2). The generation of $\pi_r$ depends on the actions in the current partial plan (the plan represented by $\mathcal{A}$) in two ways:

- The actions in the current plan are used to define an initial state for the problems of achieving the preconditions of $a$ and those in $\Sigma$. In particular, the relaxed subplan for the preconditions of $a$ is computed from the state $INIT_l$ obtained by applying the actions in $\mathcal{A}$ up to level $l-1$, ordered according to their corresponding levels.[10] The relaxed subplan for achieving $\Sigma$ is computed from $INIT_l$ modified by the effects of $a$, and it can reuse the actions in the relaxed subplan previously computed for the preconditions of $a$.

- In the process of deriving a relaxed plan, when we choose an action, we consider its potential interference with the no-ops that support a precondition of some action in $\mathcal{A}$ at a level following $l$, and we prefer actions that do not block the propagation of such no-ops. The motivation is that taking these interferences into account during the construction of a relaxed plan can lead to a better estimate of the cost required to support preconditions of type (1) and (2) in the context of the search that we are conducting to transform the current action graph into a solution graph (more details below).

---

10. Notice that, as we pointed out in the previous section, the levels in a TA-graph correspond to a total order of the actions of the represented partial-order plan that is consistent with the ordering constraints in $\Omega$ (though, of course, this is not necessarily the only valid total order).

We indicate with $Threats(a)$ the set of preconditions of the actions in $\mathcal{A}$ that would become unsupported when adding $a$ (similarly for the action preconditions that could be subverted by an action in the relaxed plan). Using the causal-link notation of partial-order planners (e.g., McAllester & Rosenblitt, 1991; Penberthy & Weld, 1992), $Threats(a)$ can be formally defined in the following way

$$Threats(a) = \{f \mid \text{no-op}(f) \text{ and } a \text{ are mutex}; \exists b, c \in \mathcal{A} \text{ such that } b \xrightarrow{f} c\}.$$

Note that, according to our representation, $b \xrightarrow{f} c$ implies $Level(b) < Level(a) < Level(c)$, where $Level(x)$ denotes the level of $x$ in $\mathcal{A}$.

Figure 3 gives a recursive algorithm for computing our relaxed plans, RelaxedPlan, which uses the following additional notation. $Duration(a)$ denotes the duration of $a$;[11] $Pre(a)$ denotes the precondition nodes of $a$; $Add(a)$ denotes the (positive) effect nodes of $a$; $Supported\_facts(l)$ denotes the set of positive facts that are true after executing the actions at levels that precede $l$ (ordered according to their level); $Num\_acts(p, l)$ denotes an estimated minimum number of actions required to reach $p$ from $Supported\_facts(l)$ (if $p$ is not reachable, $Num\_acts(p, l)$ is a negative number). The technique for computing $Num\_acts$ is described in Section 3.3.

Given a set $G$ of goal facts, an initial state $INIT_l$, and a possibly empty set of actions $A$, RelaxedPlan computes a pair $Rplan = \langle ACTS, t \rangle$ where: $ACTS$ is a set of actions including $A$ and forming a relaxed plan achieving $G$ from $INIT_l$; $t$ is a temporal value estimating the earliest time when all facts in $G$ are achieved. The first element of $Rplan$ is indicated with $Aset(Rplan)$, the second with $End\_time(Rplan)$.

As mentioned above and described in detail in Section 3.2.3, when we evaluate the addition of an action $a$, RelaxedPlan is run twice: first to compute a relaxed plan for the preconditions of $a$, and then to extend this plan for achieving the preconditions that would be subverted by $a$ (i.e., $Threats(a)$). The input set $A$ is the set of actions currently in the relaxed plan that can be "reused" to achieve an action precondition or goal of the relaxed (sub)problem. $A$ is not empty whenever RelaxedPlan is recursively executed, and when it is run to achieve $Threats(a)$.

RelaxedPlan constructs $Rplan$ through a backward process where $Bestaction(g)$ is the action $a'$ chosen to achieve a (sub)goal $g$, and such that: (i) $g$ is an effect of $a'$; (ii) all preconditions of $a'$ are reachable from $INIT_l$; (iii) the reachability of the preconditions of $a'$ requires a minimum number of actions, estimated as the maximum of the heuristic minimum number of actions required to support each precondition $p$ of $a'$ from $INIT_l$ (i.e., the maximum of $Num\_acts(p, l)$ over each precondition $p$ of $a'$); (iv) $a'$ subverts a minimum number of supported precondition nodes in $\mathcal{A}$ (i.e., the size of the set $Threats(a')$ is minimal). More formally,

$$Bestaction(g) = \underset{\{a' \in A_g\}}{ARGMIN} \left\{ \underset{p \in Pre(a')-F}{MAX} Num\_acts(p, l) + |Threats(a')| \right\},$$

where $F$ is the set of positive effects of the actions currently in $ACTS$, and $A_g$ is the set of actions with the effect $g$ and with reachable preconditions, i.e.,

---

11. If the duration of $a$ is dynamic (it depends on the value of one or more numerical variables), it is computed using the values of the numerical variables in $INIT_l$.

RelaxedPlan$(G, INIT_l, A)$

*Input*: A set of goal facts ($G$), the set of facts that are true after executing the actions of the current TA-graph up to the level $l$ ($INIT_l$), a possibly empty set of actions ($A$);

*Output*: A set of actions and a real number, estimating a minimal set of actions required to achieve $G$ and the earliest time when all facts in $G$ can be achieved, respectively.

1.  $t \leftarrow \underset{g \in G \cap INIT_l}{MAX} Time(g)$;
2.  $G \leftarrow G - INIT_l$; $ACTS \leftarrow A$;
3.  $F \leftarrow \bigcup_{a \in ACTS} Add(a)$;
4.  $t \leftarrow MAX \left\{ t, \underset{g \in G \cap F}{MAX} T(g) \right\}$;
5.  **while** $G - F \neq \emptyset$
6.      $g \leftarrow$ a fact in $G - F$;
7.      $bestact \leftarrow Bestaction(g)$;
8.      $Rplan \leftarrow$ RelaxedPlan$(Pre(bestact), INIT_l, ACTS)$;
9.      **forall** $f \in Add(bestact) - F$
10.         $T(f) \leftarrow End\_time(Rplan) + Duration(bestact)$;
11.     $ACTS \leftarrow Aset(Rplan) \cup \{bestact\}$;
12.     $F \leftarrow \bigcup_{a \in ACTS} Add(a)$;
13.     $t \leftarrow MAX\{t, End\_time(Rplan) + Duration(bestact)\}$;
14. **return** $\langle ACTS, t \rangle$.

Figure 3: Algorithm for computing a relaxed plan achieving a set of action preconditions from the state $INIT_l$. $Rplan$ is a pair of values $\langle Aset(Rplan), End\_time(Rplan)\rangle$, where the first value is a set of actions and the second is a temporal quantity. $Bestaction(g)$ is the action that is heuristically chosen to support $g$ as described in the text.

$$A_g = \{a \in \mathcal{O} \,|\, g \in Add(a), \mathcal{O} \text{ is the set of all actions}, \forall p \in Pre(a) \; Num\_acts(p) \geq 0\}.[12]$$

Notice that the set of actions $\mathcal{O}$ in the definition of $A_g$ does not contain operator instances with mutually exclusive preconditions. The reason why $Bestaction(g)$ considers the cost of the preconditions in $Pre(a') - F$, instead of in $Pre(a')$, is that the preconditions of $a'$ that are in $F$ are already supported by other actions currently in the relaxed plan under construction.

Requirements (i) and (ii) for the definition of $Bestaction$ are obvious. Regarding (iii) and (iv), we considered alternative versions that are implemented in LPG, but that are not used as default strategies because we experimentally found that on average they lead to a worse performance. In particular, instead of using the maximum of the heuristic minimum number of actions required to support each precondition $a'$, we tested the use of the *sum* of such numbers, which can give an overestimation of the actual search cost. We have also tested a version of *Bestaction* which does not consider (iv), i.e., without the term $|Threats(a')|$. While this simplified version is faster to compute, overall the performance of the planner was on average worse both in terms of CPU-time and quality of the plans produced (detailed results of this experiment are available from the web page of LPG).

---

12. In principle $A_g$ can be empty because $g$ might not be reachable from $INIT_l$ (i.e., $Bestaction(g) = \emptyset$). RelaxedPlan treats this special case by forcing its termination and returning a set of actions including a special action with a very high cost, leading $E$ to consider the element of the neighborhood under evaluation a bad possible next search state. For clarity we omit these details from the formal description of the algorithm.

Steps 1, 4 and 13 of RelaxedPlan estimate the earliest time required to achieve all goals in $G$. This is recursively defined as the maximum of

(a) the times assigned to the facts in $G$ that are already true in the state $INIT_l$ (step 1);

(b) the estimated earliest time $T(g)$ required to achieve every fact $g$ in $G$ that is an effect of an action currently in $ACTS$ (step 4);

(c) the estimated earliest time required to complete the execution of the actions chosen by *Bestaction* to achieve each of the remaining facts in $G$ (step 13).

The $T$-times of (b) are computed by steps 9–10 from the relaxed subplan derived to achieve them. Clearly the algorithm terminates, because either every (sub)goal $p$ is reachable from $INIT_l$ (i.e., $Num\_acts(p, l) \geq 0$), or at some point $bestact = \emptyset$ holds, forcing immediate termination (see footnote 12). Moreover, it can be proved that the complexity of the algorithm is polynomial in the number of actions and facts in the planning problem/domain.

### 3.2.2 AN EXAMPLE ILLUSTRATING RelaxedPlan

Suppose we are evaluating the addition of $a$ to the current TA-graph $\mathcal{A}$ illustrated in Figure 4. For each fact that is used in the example, the tables of Figure 4 give the relative $Num\_acts$-value or the temporal value ($Num\_acts$ for the unsupported facts, $Time$ for the other nodes). The $Num\_acts$-value for a fact belonging to $INIT_l$ is zero. The duration of the actions used in the example are indicated in the corresponding table of Figure 4. Solid circle and square nodes represent precondition and action nodes in $\mathcal{A} \cup \{a\}$; dotted circle and square nodes represent the precondition and action nodes that are considered during the evaluation process; finally, the gray circle and square nodes represent the precondition and action nodes that are selected by RelaxedPlan.

First we describe the derivation of the sets of actions in the relaxed plan for $Pre(a)$ and $Threats(a)$, i.e.,

$S_1 = Aset(\text{RelaxedPlan}(Pre(a), INIT_l, \emptyset))$ and
$S_2 = Aset(\text{RelaxedPlan}(Threats(a), INIT_l, S_1))$

respectively. Then we describe the derivation of the estimation of the earliest time when all preconditions in $Pre(a)$ can be achieved, i.e., $End\_time(\text{RelaxedPlan}(Pre(a), INIT_l, \emptyset))$.

*Actions for $Pre(a)$ in the Relaxed Plan*

$Pre(a)$ is $\{p_1, p_2, p_3\}$ but, since $p_2 \in INIT_l$, in the first execution of RelaxedPlan step 2 removes $p_2$ from $G$. So, only $p_1$ and $p_3$ are the goals of the relaxed problem. Suppose that in order to achieve $p_1$ we can use $a_1$, $a_2$ or $a_3$ (forming the set $A_g$ examined by $Bestaction(p_1)$ in step 7). Each of these actions is evaluated, and $a_1$ is chosen. In the recursive call of RelaxedPlan applied to the preconditions of $a_1$, $p_5$ is not considered because it already belongs to $INIT_l$. Regarding the other precondition of $a_1$ ($p_4$), suppose that $a_4$ is the only action achieving it. Then this action is chosen to achieve $p_4$, and since its preconditions belong to $INIT_l$, they are not evaluated (the new recursive call of RelaxedPlan returns an empty action set).

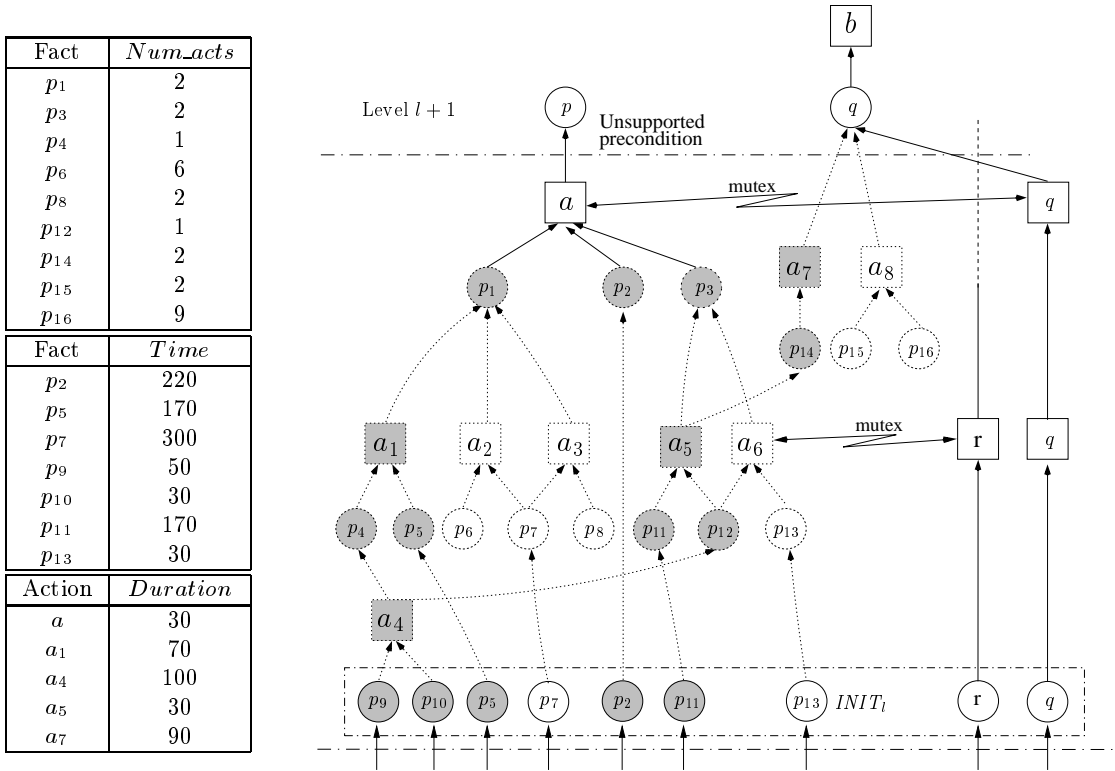

| Fact | $Num\_acts$ |
|---|---|
| $p_1$ | 2 |
| $p_3$ | 2 |
| $p_4$ | |
| $p_6$ | 6 |
| $p_8$ | 2 |
| $p_{12}$ | 1 |
| $p_{14}$ | 2 |
| $p_{15}$ | 2 |
| $p_{16}$ | 9 |

| Fact | $Time$ |
|---|---|
| $p_2$ | 220 |
| $p_5$ | 170 |
| $p_7$ | 300 |
| $p_9$ | 50 |
| $p_{10}$ | 30 |
| $p_{11}$ | 170 |
| $p_{13}$ | 30 |

| Action | $Duration$ |
|---|---|
| $a$ | 30 |
| $a_1$ | 70 |
| $a_4$ | 100 |
| $a_5$ | 30 |
| $a_7$ | 90 |

Figure 4: An example illustrating RelaxedPlan. Square nodes represent action nodes, while the other nodes represent fact nodes; solid nodes correspond to nodes in $\mathcal{A} \cup \{a\}$; dotted nodes correspond to the precondition and action nodes that are considered during the evaluation process; the gray nodes are those selected by RelaxedPlan.

Regarding the precondition $p_3$ of $a$, assume that it can be achieved only by $a_5$ and $a_6$. These actions have a common precondition ($p_{12}$) that is an effect of $a_4$, an action belonging to $ACTS$ (because it was already selected by RelaxedPlan($Pre(a_1), INIT_l, \emptyset$)). The other preconditions of these actions belong to $INIT_l$. Since $|Threats(a_5)| = 0$ and $|Threats(a_6)| = 1$, $Bestaction(p_3)$ is $a_5$. Consequently, $Aset(\mathsf{RelaxedPlan}(Pre(a), INIT_l, \emptyset))$ is $\{a_1, a_4, a_5\}$.

### Actions for $Threats(a)$ in the Relaxed Plan

Concerning the execution of RelaxedPlan for $Threats(a)$, i.e., RelaxedPlan($\{q\}, INIT_l, \{a_1, a_4, a_5\}$), suppose that the only actions for achieving $q$ are $a_7$ and $a_8$. Since the precondition $p_{14}$ of $a_7$ is an effect of $a_5$, which is an action in the input set $A$ (it belongs to the relaxed subplan computed for the preconditions of $a$), and $Threats(a_7)$ is empty, the best action chosen by RelaxedPlan to support $q$ is $a_7$. It follows that the set of actions returned by RelaxedPlan is $\{a_1, a_4, a_5, a_7\}$.

*Temporal Value for $Pre(a)$*

We now consider the evaluation of the temporal value returned by RelaxedPlan($Pre(a)$, $INIT_l, \emptyset$). According to the temporal values specified in the table of Figure 4, the value of $t$ at step 1 is $Time(p_2) = 220$. As illustrated above, RelaxedPlan for $Pre(a)$ is recursively executed to evaluate the preconditions of $a_1$ (the action chosen to achieve $p_1$) and then of $a_4$ (the action chosen to achieve $p_4$). In the evaluation of the preconditions of $a_4$, at step 1 of RelaxedPlan($Pre(a_4), INIT_l, \emptyset$) $t$ is set to 50, i.e., the maximum value between $Time(p_9)$ and $Time(p_{10})$ and the algorithm returns $\langle \emptyset, 50 \rangle$.

In the evaluation of the preconditions of $a_1$, at step 1 of RelaxedPlan($Pre(a_1), INIT_l, \emptyset$) $t$ is set to $Time(p_5) = 170$, at step 8 RelaxedPlan($Pre(a_4), INIT_l, \emptyset$)) returns $\langle \emptyset, 50 \rangle$, while steps 9–10 set $T(p_{12})$ to 50+100 (the duration of $a_4$), and at step 13 $t$ is set to $MAX\{170, 50 + 100\}$. Hence, the recursive execution of RelaxedPlan applied to the preconditions of $a_1$ returns $\langle \{a_4\}, 170 \rangle$, and at step 13 of RelaxedPlan($Pre(a), INIT_l, \emptyset$) $t$ is set to $MAX\{220, 170 + 70\} = 240$.

As we have seen in the first part of the example, the action chosen to support $p_3$ is $a_5$. The recursive execution of RelaxedPlan($Pre(a_5), INIT_l, \{a_1, a_4\}$) applied to the preconditions of $a_5$ returns $\langle \{a_1, a_4\}, 170 \rangle$. In fact, the only precondition of $a_5$ that is not in $INIT_l$ ($p_{12}$) is achieved by an action already in $ACTS$ ($a_4$). Moreover, since $T(p_{12}) = 150$ and $Time(p_{11}) = 170$, the estimated end time of $a_5$ is $170 + 30 = 200$. At step 13 of RelaxedPlan($Pre(a), INIT_l, \emptyset$) $t$ is then set to $MAX\{240, 200\}$ and the output of RelaxedPlan is $\langle \{a_1, a_4, a_5\}, 240 \rangle$.

### 3.2.3 ESTIMATING THE TERMS OF $E$

As noted before, the terms of the action evaluation function $E$ are computed by using the relaxed (sub)plan $\pi_r$ for a set of preconditions. The number of actions in $\pi_r$ and the threats of these actions are used to define a heuristic estimate of the additional search cost that would be introduced by adding an action $a$ to the current TA-graph, or removing it (i.e., the $Search\_cost$ terms of $E$). Note that in general this is not an admissible heuristic, because it can overestimate the minimum number of search steps needed to cope with the inconsistency under consideration.

The $Temporal\_cost$ term of $E(a)^i$ is an estimation of the earliest time when the new action $a$ would terminate, given the actions in $\pi_r$ and the earliest time when $\pi_r$ can be applied in the context of the current action graph.[13] The $Temporal\_cost$ term of $E(a)^r$ is an estimation of the earliest time when all preconditions that would become unsupported by removing $a$ from the current action graph could be supported again.

The $Execution\_cost$ term of $E(a)^i$ is an estimation of the additional execution cost that would be required to satisfy the preconditions of $a$, and is derived by summing the cost of each action $a'$ in $\pi_r$ ($Cost(a')$). The $Execution\_cost$ term of $E(a)^r$ is estimated similarly by

---

13. The makespan of $\pi_r$ is not a lower bound for $Temporal\_cost(a)$ because the possible parallelization of $\pi_r$ with the actions already in $\mathcal{A}$ is not considered.

EvalAdd($a$)

  *Input*: An action node $a$ that does not belong to the current TA-graph.

  *Output*: A pair formed by a set of actions and a temporal value $t$.

1.  $INIT_l \leftarrow Supported\_facts(Level(a))$;
2.  $Rplan \leftarrow$ RelaxedPlan($Pre(a), INIT_l, \emptyset$);
3.  $t_1 \leftarrow MAX\{0, MAX\{Time(a') \mid \Omega \models a' \prec a\}\}$;
4.  $t_2 \leftarrow MAX\{t_1, End\_time(Rplan)\}$;
5.  $A \leftarrow Aset(Rplan) \cup \{a\}$;
6.  $Rplan \leftarrow$ RelaxedPlan($Threats(a), INIT_l - Threats(a), A$);
7.  **return** $\langle Aset(Rplan), t_2 + Duration(a) \rangle$.

EvalDel($a$)

  *Input*: An action node $a$ that belongs to the current TA-graph.

  *Output*: A pair formed by a set of actions and a temporal value $t$.

1.  $INIT_l \leftarrow Supported\_facts(Level(a))$;
2.  $Rplan \leftarrow$ RelaxedPlan($Unsup\_facts(a), INIT_l, \emptyset$).
3.  **return** $Rplan$.

Figure 5: Algorithms for estimating the search, execution and temporal costs for the insertion (EvalAdd) and removal (EvalDel) of an action node $a$. $Rplan$ is a pair of values, identified by $Aset(Rplan)$ and $End\_time(Rplan)$, where the first is a set of actions and the second a temporal value. $Num\_acts$, $Supported\_facts$, $Duration$ and $Threats$ have been defined in Section 3.2.1. $Unsup\_facts(a)$ denotes the set of precondition nodes that become unsupported by removing $a$ from $\mathcal{A}$.

considering the preconditions of the actions that would become unsupported when removing $a$ from the current action graph. More formally, $E$ is defined as follows:

$$E(a)^i \begin{cases} Execution\_cost(a)^i = \sum_{a' \in Aset(\text{EvalAdd}(a))} Cost(a') \\ Temporal\_cost(a)^i = End\_time(\text{EvalAdd}(a)) \\ Search\_cost(a)^i = |Aset(\text{EvalAdd}(a))| + \sum_{a' \in Aset(\text{EvalAdd}(a))} |Threats(a')| \end{cases}$$

$$E(a)^r \begin{cases} Execution\_cost(a)^r = \sum_{a' \in Aset(\text{EvalDel}(a))} Cost(a') - Cost(a) \\ Temporal\_cost(a)^r = End\_time(\text{EvalDel}(a)) \\ Search\_cost(a)^r = |Aset(\text{EvalDel}(a))| + \sum_{a' \in Aset(\text{EvalDel}(a))} |Threats(a')| \end{cases}$$

where EvalAdd($a$) and EvalDel($a$) are the functions defined in Figure 5. EvalAdd($a$) returns two values: the set of actions in $\pi_r$ ($Aset$) and an estimation of the earliest time when the new action $a$ would terminate ($End\_time$). Similarly for EvalDel($a$), which returns the set of actions in the relaxed plan achieving the preconditions that would become unsupported if $a$ were removed from $\mathcal{A}$, together with an estimation of the earliest time when all these preconditions would become supported. The relaxed subplans used in EvalAdd($a$) and EvalDel($a$) are computed by RelaxedPlan, as described in Section 3.2.1.

After having computed the state $INIT_l$ using $Supported\_facts(l)$, in step 2 EvalAdd uses RelaxedPlan to compute a relaxed subplan ($Rplan$) for achieving the preconditions of the new action $a$ from $INIT_l$. Steps 3–4 compute an estimation of the earliest time when $a$ can be executed as the maximum value over the end times of all the actions preceding $a$ in $\mathcal{A}$ ($t_1$) and $End\_time(Rplan)$ ($t_2$). Steps 5–6 compute a relaxed plan for $Threats(a)$ taking account of $a$ and the actions in the first relaxed subplan.

EvalDel is simpler than EvalAdd, because the only new inconsistencies that can be generated by removing $a$ are the precondition nodes supported by $a$ (possibly through the no-op propagation of its effects) that would become unsupported. $Unsup\_facts(a)$ denotes the set of these nodes.

Of course, an action elimination from $\mathcal{A}$ to cope with an inconsistency could remove some additional inconsistencies (the unsupported preconditions of the eliminated action). Similarly, an action that is added to $\mathcal{A}$ to support a certain precondition could support additional preconditions as well. However note that, as described in Section 3.1, Walkplan, like Walksat, does not consider possible improvements during the evaluation of the search neighborhood. Hence, EvalDel and EvalAdd do not take account of additional inconsistencies that are removed from $\mathcal{A}$ as positive "side-effects" of coping with the inconsistency under consideration.

In order to illustrate the steps of EvalAdd, consider again the example of Figure 4. As shown in Section 3.2.2, the pair assigned to $Rplan$ by step 2 of EvalAdd(a) is $\langle\{a_1, a_4, a_5\}, 240\rangle$ (which is the pair of values returned by RelaxedPlan($Pre(a), INIT_l, \emptyset$)). At step 3 of EvalAdd(a) suppose that $t_1$ is set to 230 (i.e., that the highest temporal value assigned to the actions in the TA-graph that must precede $a$ is 230). Step 4 sets $t_2$ to $MAX\{230, 240\}$, and the execution of RelaxedPlan($\{q\}, INIT_l - \{q\}, \{a_1, a_4, a_5, a\}$) at step 6 returns $\langle\{a_1, a_4, a_5, a, a_7\}, t_q\rangle$, where $t_q$ is a temporal value that is ignored in the rest of the algorithm, because it does not affect the estimated end time of $a$. Thus, since the duration of $a$ is 30, the output of EvalAdd(a) is $\langle\{a_1, a_4, a_5, a, a_7\}, 240 + 30\rangle$.

## 3.3 Computing Reachability and Temporal Information

The techniques described in the previous subsection for computing the action evaluation function use heuristic reachability information about the minimum number of actions required to achieve a fact $f$ from $INIT_l$ ($Num\_acts(f, l)$), and the earliest times for actions and preconditions. LPG precomputes $Num\_acts(f, l)$ for $l = 1$ and any fact $f$, i.e., it estimates the minimum number of actions required to achieve $f$ from the initial state $I$ of the planning problem before starting the search. For $l > 1$, $Num\_acts(f, l)$ can be computed only during search because it depends on which actions nodes are in the current TA-graph (at levels preceding $l$). Since during search many action nodes can be added and removed, it is important that the computation of $Num\_acts(f, l)$ is fast.

Figure 6 gives ComputeReachabilityInformation, the algorithm used by LPG for computing $Num\_acts(f, 1)$ trying to take account of the tradeoff between quality of the estimation

ComputeReachabilityInformation$(I, \mathcal{O})$

   *Input*: The initial state of the planning problem under consideration ($I$) and all ground
       instances of the operators ($\mathcal{O}$);

   *Output*: An estimate of the number of actions ($Num\_acts$) and of the earliest time
       ($Time\_fact$) required to achieve each fact from $I$.

1.    **forall** facts $f$   /* the set of all facts is precomputed by the operator instantiation phase */
2.        **if** $f \in I$ **then**
3.            $Num\_acts(f, 1) \leftarrow 0;\ Time\_fact(f, 1) \leftarrow 0;\ Action(f, 1) \leftarrow a_{start};$
4.        **else** $Num\_acts(f, 1) \leftarrow -1;$
5.    $F \leftarrow I;\ F_{new} \leftarrow I;\ A \leftarrow \mathcal{O};$
6.    **while** $F_{new} \neq \emptyset$
7.        $F \leftarrow F \cup F_{new};\ F_{new} \leftarrow \emptyset$
8.        **while** $A' = \{a \in A \mid Pre(a) \subseteq F\}$ is not empty
9.            $a \leftarrow$ an action in $A';$
10.            $ra \leftarrow$ RequiredActions$(I, Pre(a));$
11.            $t \leftarrow \underset{f \in Pre(a)}{MAX}\ Time\_fact(f, 1);$
12.            **forall** $f \in Add(a)$
13.                **if** $f \notin F \cup F_{new}$ or $Time\_fact(f, 1) > (t + Duration(a))$ **then**
14.                    $Time\_fact(f, 1) \leftarrow t + Duration(a);$
15.                **if** $f \notin F \cup F_{new}$ or $Num\_acts(f, 1) > (ra + 1)$ **then**
16.                    $Num\_acts(f, 1) \leftarrow ra + 1;$
17.                    $Action(f, 1) \leftarrow a;$
18.            $F_{new} \leftarrow F_{new} \cup Add(a) - F;$
19.            $A \leftarrow A - \{a\};$

RequiredActions$(I, G)$

   *Input*: A set of facts $I$ and a set of action preconditions $G$;

   *Output*: An estimate of the minimum number of actions required to achieve all facts in $G$
       from $I$ ($ACTS$).

1.    $ACTS \leftarrow \emptyset;$
2.    $G \leftarrow G - I;$
3.    **while** $G \neq \emptyset$
4.        $g \leftarrow$ an element of $G;$
5.        $a \leftarrow Action(g, 1);$
6.        $ACTS \leftarrow ACTS \cup \{a\};$
7.        $G \leftarrow G \cup Pre(a) - I - \bigcup_{b \in ACTS} Add(b);$
8.    **return**$(|ACTS|).$

Figure 6: Algorithms for computing heuristic information about the reachability of each
       fact.

and computational effort to derive it. The same algorithm could be used for (re)computing $Num\_acts(f, l)$ after an action insertion/removal for any $l > 1$ (when $l > 1$, instead of $I$, in input the algorithm has $Supported\_facts(l)$).[14] In addition to $Num\_acts(f, 1)$, ComputeReachabilityInformation derives heuristic information about the possible earliest time of every fact $f$ reachable from $I$ ($Time\_fact(f, 1)$). LPG can use $Time\_fact(f, 1)$ to assign an initial temporal value ($Time(f)$) to any unsupported fact node representing $f$, instead of leaving $Time(f)$ undefined as we indicated in Section 2.2. This can give a more accurate estimation of the earliest start time of an action with unsupported preconditions, which is defined as the maximum value over the times assigned to its preconditions. Note that $Time\_fact(f, 1)$ is not updated when actions are added to (or removed from) the current TA-graph.

Before illustrating in detail the algorithm for computing reachability information that we used for the competition version of LPG, we should also note that preconditions involving numerical quantities are ignored by this technique. A new version taking numerical preconditions into account is under development.

### 3.3.1 COMPUTATION OF $Num\_acts$ AND $Time\_facts$

For clarity we first describe only the steps of ComputeReachabilityInformation used to derive $Num\_acts$, and then we comment on the computation of $Time\_fact$. In steps 1–4, the algorithm initializes $Num\_acts(f, 1)$ to 0, if $f \in I$, and to -1 otherwise (indicating that $f$ is not reachable). Then in steps 5–19 it iteratively constructs the set $F$ of facts that are reachable from $I$, starting with $F = I$, and terminating when $F$ cannot be further extended. In this forward process each action is applied at most once, when its preconditions are contained in the current $F$. The set $A$ of the available actions is initialized to the set of all possible actions (step 5), and it is reduced after each action application (step 19). The internal while-loop (steps 8–19) applies the actions in $A$ to the current $F$, possibly deriving a new set of facts $F_{new}$ in step 18. If $F_{new}$ is not empty, $F$ is extended with $F_{new}$ and the internal loop is repeated. Since $F$ monotonically increases and the number of facts is finite, termination is guaranteed. When an action $a$ in $A'$ (the subset of actions currently in $A$ that are applicable to $F$) is applied, the reachability information for its effects are revised as follows. First we estimate the minimum number $ra$ of actions required to achieve $Pre(a)$ from $I$ using the subroutine RequiredActions (step 10). Then we use $ra$ to possibly update $Num\_acts(f, 1)$ for any effect $f$ of $a$ (steps 12, 15–16). If the application of $a$ leads to a lower estimation for $f$, i.e., if $ra + 1$ is less than the current value of $Num\_acts(f, 1)$, then $Num\_acts(f, 1)$ is set to $ra + 1$. In addition, a data structure indicating the current best action to achieve $f$ ($Action(f, 1)$) is set to $a$ (step 17).[15]

For any fact $f$ in the initial state, the value of $Action(f, 1)$ is $a_{start}$ (step 3). RequiredActions uses $Action$ to derive $ra$ through a backward process starting from the input set of action preconditions ($G$), and ending when $G \subseteq I$. The subroutine incrementally constructs a set of actions ($ACTS$) achieving the facts in $G$ and the preconditions of the actions already selected (using $Action$). At each iteration the set $G$ is revised by adding the preconditions of the last action selected, and removing the facts belonging to $I$ or to the effects of actions already selected (step 7). Termination of RequiredActions is guaranteed because every element of $G$ is reachable from $I$.

---

14. In order to obtain better performance, for $l > 1$ LPG uses an incremental version of ComputeReachabilityInformation updating $Num\_acts(f, l)$ after each action insertion/removal. We omit the details of this version of the algorithm.

15. In the actual algorithm implemented in LPG, when we set $Action(f, 1)$, we consider also the case in which $Num\_acts(f, 1)$ is equal to $ra + 1$; if the execution cost of $a$ is lower than that cost of the current $Action(f, 1)$, or they have the same cost but the $a$ supports $f$ earlier, then $Action(f, 1)$ is revised to $a$. For clarity these details are omitted from the formal description of the algorithm.

$Time\_fact(f, 1)$ is computed in a way similar to $Num\_acts(f, 1)$. Step 3 of ComputeReachabilityInformation initializes it to 0, for any fact $f$ in the initial state. Then, at every application of an action $a$ in the forward process described above, we estimate the earliest possible time $t$ for applying $a$ as the maximum value over the times currently assigned to its preconditions (step 11). For any effect $f$ of $a$ that has not been considered yet (i.e., that is not in $F$), or that has a temporal value higher than $t$ plus the duration of $a$, steps 13–14 set $Time\_fact(f, 1)$ to this lower value (because we have found a shorter relaxed plan to achieve $f$ from $I$).

The complexity of ComputeReachabilityInformation is polynomial in the number of facts and actions in the problem/domain under consideration. Step 10, the most expensive step of the algorithm, is executed $O(|\mathcal{O}|)$ times, where $\mathcal{O}$ is the set of all actions, and $|\mathcal{O}|$ is the size of this set. It is easy to see that the worst-case time complexity of RequiredActions is $O(|\mathcal{O}|)$. It follows that the time complexity of ComputeReachabilityInformation is $O(|\mathcal{O}|^2)$. However, we have experimentally observed that very often RequiredActions terminates returning numbers much smaller than $|\mathcal{O}|$ (i.e, that the number of iterations that the algorithm performs is well below $|\mathcal{O}|$). Finally, we observe that the order in which actions are examined for their application in the forward process can affect the output results. In our current implementation we use a random order.

Figure 7 illustrates the algorithm with an example. Suppose that the facts in the initial state $I$ are $f_{1\ldots8}$, and that the actions in $\mathcal{O}$ are $a_{1\ldots7}$, where the subscript of the actions correspond to the order in which they are applied by the algorithm. The first actions that are applied are $a_1$, $a_2$ and $a_3$, because their preconditions are in $F$ which is initially set to $I$. The $Num\_acts$ value of these preconditions is set to zero, because RequiredActions applied to them returns zero. In the internal for-loop of the algorithm we update the reachability information for each effect of these actions. In particular, consider the effects $f_1$ and $f_9$ of $a_1$. Since $f_1$ is not a new fact (it belongs to $I$) and its $Num\_acts$ and $Time\_fact$ values are set to the minimum (initial) values, steps 14 and 16 do not revise them. Since $f_9$ is a new fact, step 14 sets $Time\_fact(f_9)$ to 10 (i.e., the duration of $a_1$), and step 16 sets $Num\_acts(f_9)$ to 1 ($ra$ is zero). Moreover, $Action(f_9, 1)$ is set to $a_1$ by step 17. The effects of $a_2$ and $a_3$ are handled similarly.

At this point, since there is no other action that is applicable in $F$, the internal while-loop terminates, $F$ is set to $F \cup \{f_9, f_{10}, f_{11}, f_{12}\}$, and $F_{new}$ is set to $\emptyset$. The set $A'$ of the actions in $A$ that are applicable is $\{a_4, a_5, a_6\}$. Consider the application of $a_4$. We have that $ra$ at step 10 is set to 2, because RequiredActions$(I, Pre(a_4))$ sets $ACTS$ to $\{a_1, a_2\}$ (note that $f_1 \in I$, $Action(f_9, 1) = a_1$, $Action(f_{10}, 1) = a_2$, and all preconditions of these actions are in $I$). Thus, $Time\_fact(f_{13})$ is set to 80 (i.e., the maximum temporal value assigned to a precondition of $a_4$, 30, plus the duration of $a_4$), and $Num\_acts(f_{13})$ to 3. The effects of the actions $a_5$ and $a_6$ are handled in a similar way. However, it is worth noting that $Num\_acts(f_{15})$ is first set to 3, when we examine $a_5$, and then revised to 2, when we examine $a_6$. Analogously, $Time\_fact(f_{15})$ is first set to 120, and then revised to 80, while $Action(f_{15}, 1)$ is first set to $a_5$ and then to $a_6$.

| Actions | $ra$ | Duration |
|---------|------|----------|
| $a_1$ | 0 | 10 |
| $a_2$ | 0 | 30 |
| $a_3$ | 0 | 50 |
| $a_4$ | 2 | 50 |
| $a_5$ | 2 | 70 |
| $a_6$ | 1 | 30 |
| $a_7$ | 5 | 20 |

| Facts | $Num\_acts$ | $Time\_fact$ |
|-------|-------------|--------------|
| $f_9$ | 1 | 10 |
| $f_{10}$ | 1 | 30 |
| $f_{11}$ | 1 | 30 |
| $f_{12}$ | 1 | 50 |
| $f_{13}$ | 3 | 80 |
| $f_{14}$ | 3 | 120 |
| $f_{15}$ | 2 | 80 |
| $f_{16}$ | 2 | 80 |
| $f_{17}$ | 7 | 140 |

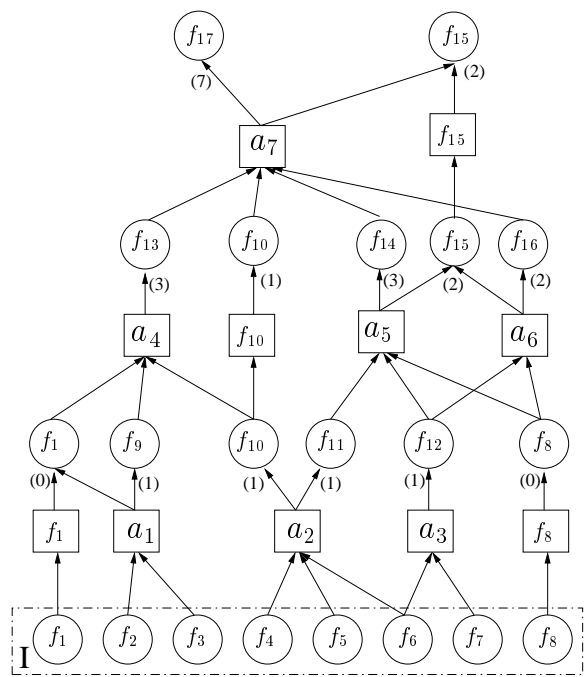

Figure 7: An example illustrating ComputeReachabilityInformation. The numbers in parenthesis are $Num\_acts$ values. $\mathcal{O} = \{a_1, a_2, ..., a_7\}$. The subscript of each action corresponds to the order in which it is applied.

Consider now the preconditions of the last applicable action $a_7$. RequiredActions applied to $Pre(a_7)$ by step 10 returns 6, because the set $ACTS$ of actions selected by the subroutine is $\{a_4, a_1, a_2, a_5, a_3, a_6\}$. Steps 11 of the ComputeReachabilityInformation sets $t$ to $Time\_fact(f_{14}) = 120$, and hence the $Time\_fact$-value for the new effect $f_{17}$ is set to 120+20, while its $Num\_acts$-value is set to 6+1.

### 3.3.2 Related work on reachability information

Other techniques for estimating the cost of reaching a fact (or a set of facts) from a certain state have been proposed and used in some planners, e.g., HSP (Bonet & Geffner, 2001), FF (Hoffmann & Nebel, 2001) and SAPA (Do & Kambhampati, 2002). When comparing ComputeReachabilityInformation with these techniques, we should note that in LPG the $Num\_acts$-values are used to select the actions forming relaxed plans achieving sets of action preconditions or goals (computed by RelaxedPlan). These relaxed plans are then used by the action evaluation function $E$ guiding the search. In general, this approach is similar to FF's and SAPA's methods, but with some significant differences that we comment on below.

Bonet & Geffner proposed two basic heuristics for HSP, $h_{max}$ and $h_{add}$. In $h_{add}$ the (search) cost of a set of facts is the sum of the costs of each individual fact, while in $h_{max}$ it is the maximum cost over all individual costs. As noted by Haslum and Geffner (2000), $h_{max}$ and $h_{add}$ are approximation of the optimal cost function of a relaxed problem where delete effects are ignored. $h_{add}$ ignores positive interactions among subgoals that could make one goal simpler after a second one has been achieved (this makes $h_{add}$ non-admissible). $h_{max}$ is admissible, but it is less informative and effective for Bonet & Geffner's HSP planner.

A difference between the forward process of our algorithm for computing $Num\_acts$ and Bonet & Geffner's forward propagation for computing $h_{add}$ is that in our propagation every action is applied at most once, while in their propagation it can be considered more than once (for computing $h_{max}$ it suffices to apply each action once with an appropriate order). This restriction, that we introduced for efficiency reasons, can clearly lead to overestimation of reachability costs. By adding a new step between steps 17 and 18 of ComputeReachability-Information that adds to $A$ every action with $f$ as precondition, we can obtain a more accurate cost propagation like in $h_{add}$. However, this could slow down the planning process, given that reachability information may be (re)computed many times during search.

Another important difference concerns the use of the subroutine RequiredActions at step 10 for estimating the cost of reaching a set of preconditions $G$. Instead of considering the maximum value over the costs of the preconditions or the sum of their costs, like in $h_{max}$ and $h_{add}$, respectively, we compute a relaxed plan for $G$, and we count the number of actions forming it. This can be seen as an intermediate approach between the $h_{max}$ and $h_{add}$ methods, aimed at taking account of positive interactions among subgoals.[16]

Finally, another difference concerns the initial set of actions used in the forward process. While our set ($\mathcal{O}$) does not contain actions with preconditions that are mutex, Bonet & Geffner's forward processes for computing $h_{max}$ and $h_{add}$ contain them. The use of our restricted set of actions would make the approximation of $h_{max}$ and $h_{add}$ more accurate.

As observed by Hoffmann (2001), FF's reachability technique is similar to $h_{max}$, and so the previous observations about $h_{max}$ compared to our reachability information hold also for FF's technique. Another difference with respect to FF concerns the choice of actions forming the relaxed plans. While RelaxedPlan and EvalAdd take threats into account, FF's relaxed plans do not consider them. Moreover, FF can generate relaxed plans including actions with mutex preconditions, while we exclude such actions.

Most of the differences with respect to HSP's and FF's reachability information that we have outlined appear to also hold when comparing ComputeReachabilityInformation and SAPA's reachability techniques (in particular, the use of RequiredActions for estimating the search cost of a *set* of preconditions, the application of an action at most once in the forward process, and the use of a more restrictive set of actions $\mathcal{O}$). Another significant difference is that, while SAPA's reachability information concerns execution and temporal costs, our information concerns mainly search costs. As a consequence, the action choices in RelaxedPlan depend mainly on the search costs (as we pointed out, when there is more than one action with the lowest search cost, $Bestaction$ chooses the one with lower execution cost). In LPG the execution and temporal costs of the relaxed plans are subsequently taken into account by the action evaluation function $E$, using the actions in the computed relaxed plans. A major motivation for giving primary importance to search costs was that we designed our planner as an any time planning system, that can compute a first solution quickly, and then derive additional solutions with incrementally better quality, but requiring more CPU-time (this incremental process is described in Section 3.6).

---

16. Note that if we replaced step 10 with $ra \leftarrow$ "sum of $Num\_acts(f, 1)$ for each $f$ in $Pre(a)$", then the resulting algorithm would be quite similar to $h_{add}$ (using the additional step for reconsidering actions already applied that we mentioned above).

### 3.4 Updating Ordering Constraints and Temporal Values

In this subsection we describe the generation during search of action ordering constraints in the current TA-graph $\mathcal{A}$, and the update at each search step of the temporal values associated with the fact and action nodes of $\mathcal{A}$. If during search the planner adds an action node $a$ to $\mathcal{A}$ for supporting a precondition of another action node $b$, then $a \prec_C b$ is added to $\Omega$. Moreover, for each action $c$ in $\mathcal{A}$ that is mutex with $a$, if $Level(a) < Level(c)$, then $a \prec_E c$ is added to $\Omega$, otherwise ($Level(c) < Level(a)$) $c \prec_E a$ is added to $\Omega$. If the planner removes $a$ from $\mathcal{A}$, then any ordering constraint involving $a$ is removed from $\Omega$.

The addition/removal of an action node $a$ also determines a possible revision of $Time(x)$, the temporal value assigned to any fact and action $x$ that is (directly or indirectly) connected to $a$ through the ordering constraints in $\Omega$. Essentially, the algorithm for revising the temporal values assigned to the nodes of $\mathcal{A}$ performs a simple forward propagation starting from the effects of $a$, and updating level by level the times of the actions (together with the relative precondition and effect nodes) that are constrained by $\Omega$ to start after the end of $a$. If every precondition is of type `at end` and every effect is of type `at end`, when an action node $a'$ is considered for possible temporal revision, $Time(a')$ becomes the maximum value over the temporal values assigned to (a) its preconditions and (b) the actions preceding $a'$ according to $\Omega$, plus the duration of $a'$. The times assigned to the effect nodes of $a'$ are revised accordingly. If $a'$ is the only action node supporting a precondition node $f$, or its temporal value is lower than the value assigned to the other action nodes supporting $f$, then $Time(f)$ is set to $Time(a')$. For instance, suppose that, in order to support the precondition node $f_7$ of $a_4$ in the TA-graph in Figure 1, we insert the action node $a_5$ at level 4 (see Figure 8). $a_5$ has duration 110 and precondition node $f_8$. Since $Time(f_8) = 120$, $Time(f_7)$ becomes 230, which is propagated to $a_4$ and its effects. $Time(a_4)$ becomes 270, $Time(f_{12})$ is revised to 220 ($Time(a_3)$), $Time(f_{13})$ is revised to 270 ($Time(a_4)$), while $Time(f_8)$ remains 120 ($Time(a_2)$).

Some operators in the domains used for the 3rd IPC contain (pre)conditions of type "`at end`" or "`at start`", and effects of type "`at start`" (Fox & Long, 2003), i.e., preconditions that must hold at the end or at the beginning of the action, and effects that are true at the beginning of the action. In the following we revise the definition of $Time$ that we have given in Section 2.2 to consider actions involving these types of preconditions/effects. When an action node $a$ has a precondition node $p$ of type `at end`, in the definition of $Time(a)$, we use $Time(p) - Duration(a)$ instead of $Time(p)$.[17] While in the definition of $Time(a)$ precondition nodes of type `at start` are treated as precondition nodes of type `overall`.

---

17. In the special cases in which a precondition of $a$ is of type either `at end` or `overall`, and it is also an effect node of type `at start` of $a$, $Time(p)$ is not considered in the definition of $Time(a)$ because the action itself makes $p$ true (unless $p$ is also a precondition node of $a$ of type `at start`).

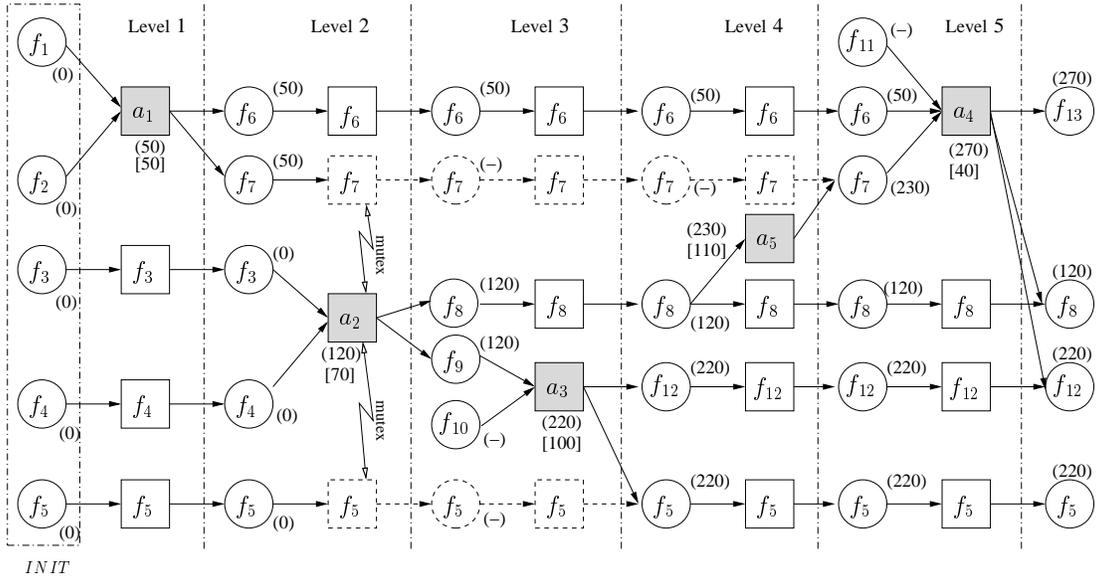

$$\Omega = \{a_1 \prec_C a_4;\ a_2 \prec_C a_3; a_2 \prec_C a_5; a_5 \prec_C a_4;\ a_1 \prec_E a_2; a_2 \prec_E a_3\ a_2 \prec_E a_4\}$$

Figure 8: Update of the TA-graph of Figure 1 after the addition of action node $a_5$ at level 4 to support the precondition node $f_7$ of $a_4$. Dashed edges form chains of no-ops that are blocked by mutex actions. Round brackets contain temporal values assigned by $\mathcal{T}$ to the fact nodes (circles) and the action nodes (squares). The numbers in square brackets represent action durations. "(−)" indicates that the corresponding fact node is not supported.

If an action node $a$ has an effect node $e$ of type $\mathtt{at\ start}$, when we estimate $Time(e)$, instead of using $Time(a)$, we use the minimum value over (1) $Time(a')$, for any action node $a'$ supporting $e$ by an effect of type $\mathtt{at\ end}$, (2) $Time(a'') - Duration(a'')$, for any action node $a''$ supporting $e$ by an effect of type $\mathtt{at\ start}$, and (3) $Time(a) - Duration(a)$ (because $e$ is supported at the start time of $a$).

When preconditions of type different from $\mathtt{over\ all}$ and effects of type different from $\mathtt{at\ end}$ are present in the domain specification, in some cases two mutex actions can partially overlap.[18] The version of LPG that took part in the 3rd IPC did not handle these possible overlaps, and any pair of mutex actions was treated by always imposing an ordering constraint between the end of an action and the start of the other one. While this is always a sound way of ordering mutex actions, it might over-constrain the actions, introducing unnecessary delays in the plan. However, in the test problems of the 3rd IPC the possibility of overlapping mutex actions is rare, and when it is possible it does not affect the temporal quality of the plans significantly. Recently, we have extended the treatment of mutex actions in LPG, distinguishing various types of interferences and competing needs. These are handled by ordering constraints between different endpoints of the involved actions, allowing overlapping mutex actions. Experimental results using the SimpleTime variant of the Satellite domain show that the new version of LPG generates plans which are about 10% better (in terms of makespan) than those computed by the competition version. Moreover, the overhead introduced by the more sophisticated management of the temporal information is on average negligible. A detailed description of how temporal information is managed in the new version of LPG is given in another recent paper (Gerevini, Saetti, & Serina, 2003).

---

18. For example, if $f$ is a precondition $\mathtt{at\ start}$ of $a$ and $\neg f$ is an effect $\mathtt{at\ end}$ of $b$, although $a$ and $b$ are mutex, $a$ can overlap $b$ (e.g., $a$ can start after the start time of $b$ and terminate before the end time of $b$). If $a$ is at a level preceding the level of $b$, the only ordering constraint that should be imposed is that *the start of $a$ is before the end of $b$*. Otherwise, the imposed ordering constraint is that the end of $b$ is before the start of $a$, and so the two actions do not overlap.

## 3.5 Numerical State Variables

In this subsection we briefly describe how LPG deals with preconditions and effects involving numerical quantities. We start with a brief description of numerical preconditions and effects in PDDL2.1, and then we show how the plan representation, search neighborhood and heuristics that we presented in the previous sections have been extended to handle them.

In PDDL2.1 a state $s$ for a planning domain involving numerical variables is a pair $\langle p(s), v(s) \rangle$ where $p(s)$ is a set of ground atoms (positive facts), and $v(s) = \langle r_1, \ldots, r_n \rangle$ is a tuple of real numbers representing the values of the $n$ numerical variables $v^1$, $v^2$, ..., $v^n$. A numerical expression is an arithmetic expression over the set $V$ of these variables and the real numbers. A numerical precondition is a triple $\langle exp, rel, exp' \rangle$, where $exp$ and $exp'$ are numerical expressions, and $rel \in \{<, \leq, =, \geq, >\}$ is a relational operator. A numerical effect is a triple $\langle v^i, ass, exp \rangle$, where $v^i \in V$ is a variable, $ass \in \{:=, +=, -=, *=, /=\}$ is an assignment operator (using a C-like notation), and $exp$ is a numerical expression.

In order to handle numerical domains, we have extended the notion of TA-graph with *numerical fact nodes* representing values of numerical variables. For each level $l$ in the current action graph $\mathcal{A}$ and each numerical variable $v^i \in V$, there is a numerical fact node representing the value for $v^i$ at level $l$. The resulting tuple of real values at level $l$ is denoted by $Num\_values(l)$. These values are derived by applying all actions in $\mathcal{A}$ at the levels preceding $l$, starting from the initial level and following the order of the corresponding levels.[19] The values of the numerical fact nodes at the initial level, $Num\_values(0)$, are the real numbers assigned to the corresponding numerical variables in the initial state of the planning problem under consideration. Similarly, we can associate with each level $l$ a set of facts that are true ($Supported\_facts(l)$), given the actions in $\mathcal{A}$ at levels preceding $l$. In this way we can define a numerical state $s_l = \langle Supported\_facts(l), Num\_values(l) \rangle$ for each level $l$ of $\mathcal{A}$.

A numerical precondition $\langle exp, rel, exp' \rangle$ of an action at a level $l$ is supported if and only if the values of $exp$ and $exp'$ evaluated in $s_l$ satisfy the relation $rel$.

---

19. This way of ordering actions at levels before $l$ is consistent with the action ordering constraints in $\mathcal{A}$ (if any). Furthermore, note that if an action $a$ at level $l$ has a numerical precondition involving a numerical variable $v^j$, then any action $b$ with an effect affecting the value of $v^j$ is mutex with $a$. So, if $\mathcal{A}$ is a TA-graph, then $b \prec_E a \in \Omega$. Otherwise ($\mathcal{A}$ is a simple LA-graph without temporal information), $b \prec a$ is implied by the levels of $a$ and $b$.

Every time an action is added/removed to/from a level of $\mathcal{A}$ we apply/retract the numerical effects of the action, which can modify the values associated with some numerical fact nodes at the next level. These changes are propagated to the following levels of the graph. During this propagation, we identify the numerical preconditions that become supported or unsupported. Moreover, if the value of a numerical fact node affecting the duration of an action is changed, then we update the duration of this action.

The local search neighborhood associated with an unsupported numerical precondition $p = \langle exp, rel, exp' \rangle$ of an action $a$ is defined as the set of linear action graphs obtained by either removing $a$, or adding a new action that *decreases* the "gap" between the values of $exp$ and $exp'$ according to $rel$ (possibly supporting $p$).[20] In the competition version of LPG, we considered adding an action only to the level immediately before the level of $p$ (while for a boolean unsupported precondition $q$ an action supporting it can be added to *any* preceding level). We are currently studying an extension of the neighborhood in which supporting actions can be added to any preceding level also in the case of numerical preconditions.

We now briefly describe how LPG computes the relaxed plans used by EvalAdd and EvalDel for numerical domains. This is done by an extended version of RelaxedPlan handling numerical preconditions in a very simple way. Since the current version of ComputeReachabilityInformation ignores numerical preconditions, there is no $Num\_acts$-value associated with them. Hence, when RelaxedPlan chooses the (heuristic) best action to support a subgoal $g$, for each numerical precondition $p$ involved in the definition of $Bestaction(g)$ (see Section 3.2.1), $Num\_acts(p, l)$ is replaced by 1, i.e., the estimated minimum cost to satisfy any numerical precondition is always 1. (Of course, this is quite a strong assumption giving weak information; we are currently working on a new version of the planner using more informative heuristics for constructing relaxed plans involving numerical preconditions.) Another difference in the definition of $Bestaction(g)$ is that, if $g$ is a numerical precondition, instead of considering only the actions supporting $g$, we consider every action that decreases the gap between the values of the expressions forming $g$.

The relaxation of the plans computed by the extended version of RelaxedPlan concerns both the negative effects, which are ignored for plan validity (but considered to count possible threats), and a form of *monotonic* change of the minimum and maximum possible values for the numerical quantities. We start from the numerical initial state $INIT_l = \langle Supported\_facts(l), Num\_values(l) \rangle$ and, for each numerical variable involved in an action in the relaxed plan constructed from $INIT_l$, we consider only the minimum/maximum values that the variable can assume given the actions already in the relaxed plan. These values are monotonically decreased/increased whenever an action is added to the relaxed plan. Specifically, we define two tuples of numerical values, $v_{max}$ and $v_{min}$, that are both initialized using $Num\_values(l)$. If an effect of an action in the relaxed plan increases the value of a variable $v^i$ by a quantity $\delta$, then we increase $v^i_{max}$ by $\delta$; while, if it decreases the value of $v^i$ by $\delta$, then we decrease $v^i_{min}$ by $\delta$. During the construction of a relaxed plan, when we check whether a numerical precondition $p = \langle v^x, >, v^y \rangle$ is supported, we evaluate $v^x > v^y$ considering $v^x_{max}$ as the value assigned to $v^x$, and $v^y_{min}$ as the value assigned to $v^y$. If an expression involves more than one numerical variable (e.g., $p = \langle v^x - v^y, >, v^z \rangle$), we consider the combination of the maximum/minimum values that is most favorable to satisfy the condition (the value of $v_x$ is $v_{max}$, the value of $v^y$ is $v_{min}^y$, and the value of $v^z$ is $v_{min}^z$). Similarly if the expression involves another relational operator.

---

20. Note that it can be necessary to add more than one action to support a numerical precondition. These actions are added to different levels by different search steps. For instance, suppose that $p = \langle x, >, 100 \rangle$ is a numerical precondition, and that $a$ is the only action with a numerical effect $e$ increasing the value of $x$. If $e$ increases $x$ by 20 and the current value of $x$ is 30, then we need four actions to support $p$.

## 3.6 Multi-Criteria Incremental Plan Quality

As we have seen, our approach can model different plan quality criteria determined by action execution costs and action durations. The coefficients $\alpha$, $\beta$ and $\gamma$ of the action evaluation function $E$ specified in Section 3.2 are used to weigh the relative importance of the execution and temporal costs of $E$, as well as to normalize them with respect to the search cost. Specifically, LPG uses the following function for evaluating the insertion of an action node $a$ (the evaluation function $E(a)^r$ for removing an action node is analogous):

$$E(a)^i = \frac{\mu_E}{max_{ET}} \cdot Execution\_cost(a)^i + \frac{\mu_T}{max_{ET}} \cdot Temporal\_cost(a)^i + \frac{1}{max_S} \cdot Search\_cost(a)^i,$$

where $\mu_E$ and $\mu_T$ are non-negative coefficients that weigh the relative importance of the execution and temporal costs, respectively. Their values can be set by the user, or they can be automatically derived from the expression defining the plan metrics in the formalization of the problem. The factors $1/max_{ET}$ and $1/max_S$ are used to normalize the terms of $E$ to a value less than or equal to 1. The value of $max_{ET}$ is defined as $\mu_E \cdot max_E + \mu_T \cdot max_T$, where $max_E$ ($max_T$) is the maximum value of the first (second) term of $E$ over all TA-graphs in the neighborhood, multiplied by the number $\kappa$ of inconsistencies in the current action graph; $max_S$ is defined as the maximum value of $Search\_cost$ over all possible action insertions/removals that eliminate the inconsistency under consideration. The role of $\kappa$ is to decrease the importance of the first two optimization terms when the current plan contains many inconsistencies, and to increase it when the search approaches a valid plan. I.e., $E(a)^i$ can be rewritten as

$$E(a)^i = \frac{1}{\kappa \cdot (\mu_E \cdot max_E + \mu_T \cdot max_T)} \cdot (\mu_E \cdot Execution\_cost(a)^i + \mu_T \cdot Temporal\_cost(a)^i) +$$
$$+ \frac{1}{max_S} \cdot Search\_cost(a)^i.$$

Without this normalization the first two terms of $E$ could be much higher than the value of the third term. This would guide the search towards good quality plans without paying sufficient attention to their validity. Instead, we would like to have the search give more importance to reducing the search cost, rather than optimizing the quality of a plan, especially when the current partial plan contains many inconsistencies,

Our planner can produce a succession of valid plans where each plan is an improvement of the previous ones in terms of quality. The first plan generated is used to initialize a new search for a second plan with better quality, and so on. This is a process that incrementally improves the quality of the plans, and the search can be stopped at any time to give the best plan computed so far (each plan can be written in a file as soon as it is derived). When LPG starts a new search, some inconsistencies are forced in the TA-graph representing the previous plan, and the resulting TA-graph is used to initialize the search. Similarly, during search some random inconsistencies are forced in the current TA-graph when a valid plan that does not improve the plan of the previous search is reached. This is done by choosing a small set $R$ of action nodes that are removed from the action graph together with (1) the action nodes supporting their preconditions and (2) the action nodes with a precondition supported by an action in $R$. The elements of $R$ are chosen by taking account of the values of $\mu_E$ and $\mu_T$. If $\mu_E > \mu_T$, we randomly remove action nodes giving higher probability to those representing actions with higher execution costs, otherwise preference is given to the action nodes having a higher impact on the plan makespan.[21]

In the 3rd IPC, for each test problem attempted, we considered only the first and the last solutions generated by LPG within five CPU-minutes. The first solution was used to test how fast our planner can be; the last solution to test how good a solution can be. Often the first solution has low quality compared to the last one, while the last solution requires much more CPU-time than the first. The other fully-automated planners in the competition did not exhibit any-time behavior like LPG. So, when we compare our two solutions with the single solution derived by the other planners, we should consider that LPG very often derives additional solutions of intermediate quality, and requiring intermediate CPU-time. In particular, as will be shown in the next section, it can be the case that, when (1) the first solution found by LPG requires less CPU-time than any other planner, but has quality worse than the best solution found by the other planners, and (2) the last solution of LPG has superior quality to all other planners but requires more CPU-time, LPG finds an intermediate solution which is still better than the solutions found by all other planners *and* is derived in less CPU-time.

## 4. Experimental Results

All our techniques are implemented in LPG. The system is written in C and is available from `http://prometeo.ing.unibs.it/lpg`. In this section we present some experimental results illustrating the efficiency of LPG using the test problems of the 3rd IPC. These problems belong to several domains, most of which have some variants containing different features of PDDL2.1. The variants are named "Strips", "SimpleTime", "Time", "Complex", "Numeric" and "HardNumeric", and are all handled by our planner. For a description of the domains and of the relative variants, the reader can visit the official web site of the 3rd IPC (`www.dur.ac.uk/d.p.long/competition.html`).

All tests were conducted on the official machine of the competition, an AMD Athlon$^{tm}$ MP 1800+ (1500MHz) with 1 Gbyte of RAM. The results for LPG correspond to median values over five runs for each problem considered. The CPU-time limit for each run was 5 minutes, after which termination was forced.[22] Notice that the results that we present here are not exactly the same as the official results of the competition, where for lack of time we were not able to run our system a sufficient number of times to obtain meaningful statistical data. However, in general the new results are very similar to those of the competition, with some considerable improvement in Satellite Complex and in the Rovers domains, where many problems could not be solved due to a minor bug in the parser of our planner that was easily fixed right after the competition.

---

21. In the version of LPG that took part in the 3rd IPC this second preference was based on a simple estimation of the temporal impact of each action node. We are currently testing a newer version that selects such actions more accurately by using the *critical path* in the graph of the ordering constraints in the TA-graph.

22. When the CPU-time limit was exceeded in one or two runs, the median values are derived by considering these runs as those producing the worst results. When the CPU-time limit was exceeded in three or four runs, instead of the median values, we considered the worst results of the remaining successful runs. This happened in 16 of the 442 problems solved.

| Planner | Solved | Attempted | Success ratio |
|---------|--------|-----------|---------------|
| LPG | 442 | 468 | 94% |
| FF | 237 | 284 | 83% |
| Simplanner | 91 | 122 | 75% |
| Sapa | 80 | 122 | 66% |
| MIPS | 331 | 508 | 65% |
| VHPOP | 122 | 224 | 54% |
| Stella | 50 | 102 | 49% |
| TP4 | 26 | 204 | 13% |
| TPSYS | 14 | 120 | 12% |
| SemSyn | 11 | 144 | 8% |

Table 1: Number of problems attempted and solved by the planners that took part in the 3rd IPC ordered by their success ratio. The data from the planners compared with LPG are from the official web site of the 3rd IPC. The data for LPG do not consider the 20 problems in Satellite HardNumeric, which are all solved by the current version of the planner, slightly improving the success ratio.

Overall, the number of problems attempted in the new tests by our planner was 468 (over a total of 508 problems), and the success ratio was 94.4% (the problems attempted by LPG in the competition were 428 and the success ratio 87%). Figure 1 gives these data for every fully-automated planner that took part in the competition. The success ratio of LPG is the highest one over all competing domain-independent planners.

The version of LPG that we used in the competition is integrated with an alternative search method that can be activated when the local search is not effective. This method is based on the same best-first search technique implemented in FF (Hoffmann & Nebel, 2001). The only domain were we used best-first search instead of local search is FreeCells. The 40 problems that were not attempted by our planner are the 20 problems in Settlers Numeric and the 20 problems in Satellite HardNumeric. The first domain contains operators with universally quantified effects, which are not handled in the current version of LPG. The plan metrics of the problems in the second domain require maximizing the value of a certain numerical variable representing acquired data (data-stored), which is another feature of PDDL2.1 that the competition version of LPG did not handle properly. Many of these problems were solved by the other fully-automated planners by the empty plan or by plans with zero quality. While such plans could have been derived also by LPG, we did not consider these interesting solutions.[23]

---

23. Very recently we have extended LPG to handle maximization of plan metric expressions. This new version of LPG solves all the 20 test problems of Satellite HardNumeric, generating plans with quality higher than zero. The only fully-automated planner of the competition that derived solutions with data-stored > 0 is MIPS. An experimental comparison of MIPS and LPG considering only these solutions

We ran LPG with the same default settings for every problem attempted (maximum numbers of search steps and restarts for each run, inconsistency selection strategy, and noise factor), that can be modified by the user. The default initial value of the noise $p$ is 0.1. Note that this is a dynamic value that is automatically increased/decreased by the planner during search, depending on the variance of the number of inconsistencies in the last $n$ search steps. In all our tests $p$ was automatically increased if the variance did not change significantly in the last 50 search steps. It was set to the initial default value otherwise. The parameters $\mu_E$ and $\mu_T$ of the action evaluation function were automatically set using the (linear) plan metric specified in the problem formalization. In particular, $\mu_E$ was set to 1, while $\mu_T$ was set to the coefficient weighing the `total-time` variable in the expression specifying the plan metric. For instance, in the example of plan metric given at the end of Section 2, the coefficient weighing `total-time` is 4, and so for that problem $\mu_T$ was set to 4. If no plan metric was specified, then $\mu_E$ was set to 0.5 and $\mu_T$ to 0.

The performance of LPG was tested in terms of both CPU-time required to find a solution (LPG-speed) and quality of the best plan computed (LPG-quality) using at most five CPU-minutes. In the plots of Figures 9, 10, 11 and 12, on the x-axis we have the problem names (simplified with numbers); on the y-axis, in the plots for CPU-time we have milliseconds (logarithmic scale), while in the plots for plan quality we have the quality of the plans generated, measured using the plan metric expression in the corresponding problem specification. Note that the lower the plan quality values, the better the corresponding plans are.

Figure 9 shows the performance of LPG-speed compared to the other competitors in some variants of four domains.[24] In `DriverLog` Strips, FF is on average the fastest planner, but LPG solves more problems, and it scales up somewhat better. In `ZenoTravel` SimpleTime, LPG outperforms the other competitors in terms of both number of problems solved and CPU-time (our planner is about one order of magnitude faster). In `Satellite` Complex the excellent performance of LPG is even more evident especially for the largest problems. Finally, in `Rovers` Numeric, FF and LPG perform similarly, but our planner solves a larger number of problems. The plots concerning the performance of LPG-quality for these four domain variants are given in Figure 10. These results show that the solution computed by our planner was always similar to or better than the solution derived by any of the other planners. The most interesting differences are in `Satellite` Complex, where LPG-quality produced solutions of higher quality for almost every problem.

In order to derive some general results about the performance of our approach with respect to all other fully-automated planners of the competition, we compared LPG with the best result over *all* these planners. We will indicate these results as if they were produced by a hypothetical "SuperPlanner" (which does not exist). Clearly, if LPG performs generally better than the SuperPlanner in a certain domain, then in that domain it performs better than *any* other real planner that we considered. On the other hand, if it performs worse,

---

shows that the plans generated by LPG have quality much higher than those computed by MIPS, and that our planner is significantly faster (detailed results of this experiment are available from the web page of LPG).

24. Complete results for all other domains and variants are available from `http://prometeo.ing.unibs.it/lpg/test-results`.

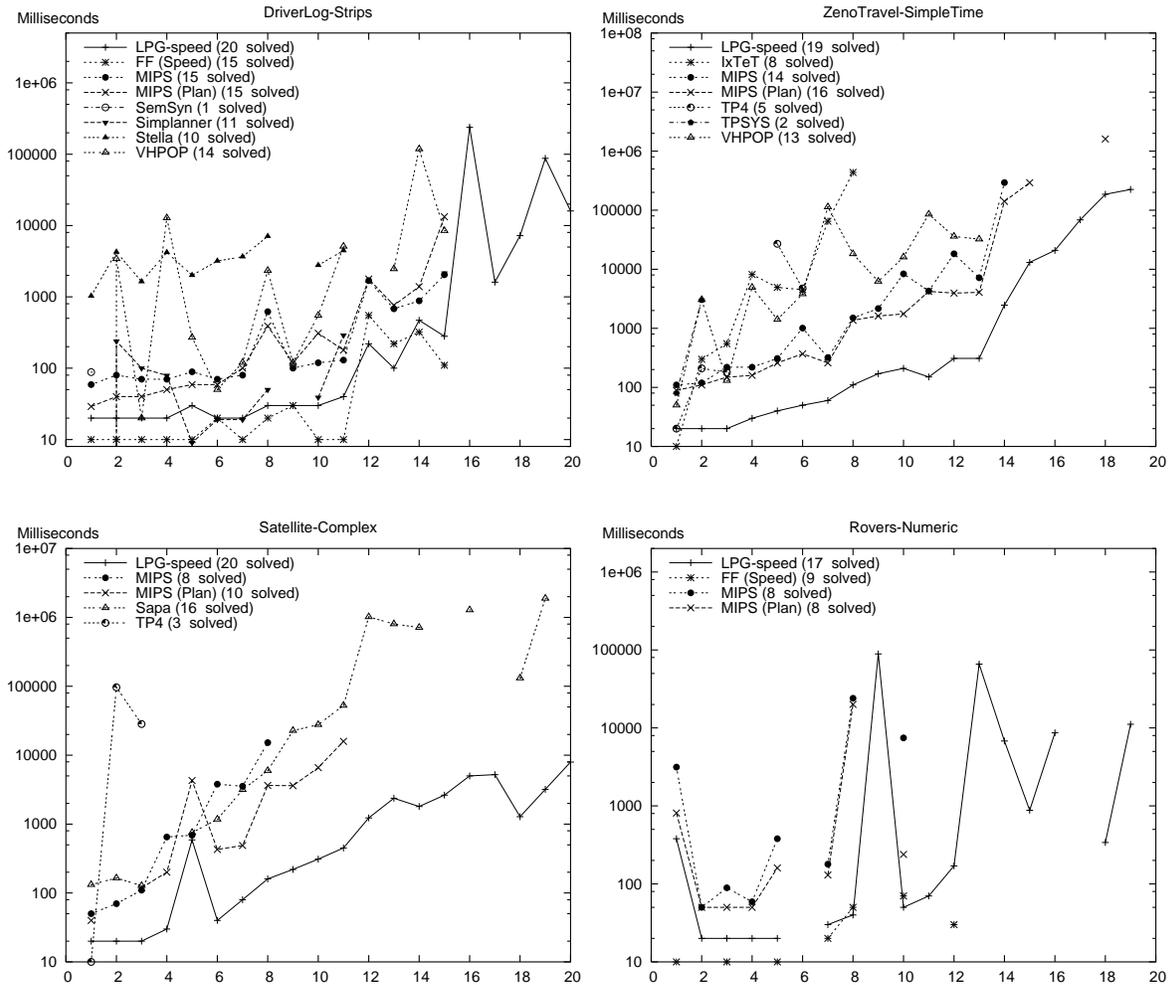

Figure 9: CPU-time and number of problems solved by the fully-automated planners of the 3rd IPC for the domains `DriverLog` Strips, `ZenoTravel` SimpleTime, `Satellite` Complex and `Rovers` Numeric.

this does not necessarily imply that there is a single real planner that generally performs better than LPG.

The plots of Figures 11 and 12 give complete results for `Satellite`, one of the domains where our planner performed particularly well in the temporal and Complex variants. The plots on the left show CPU-times for LPG-speed, LPG-quality, and the two corresponding versions of the SuperPlanner: a version in which, for each problem, we consider the fastest planner over all the other fully-automated planners, and a version in which we consider the planner that produced the best quality plan (of course it can be the case that the fastest planner for a problem is different from the planner that produces the best quality solution for that problem). The plots on the right show plan quality for the two versions of LPG and the SuperPlanner.

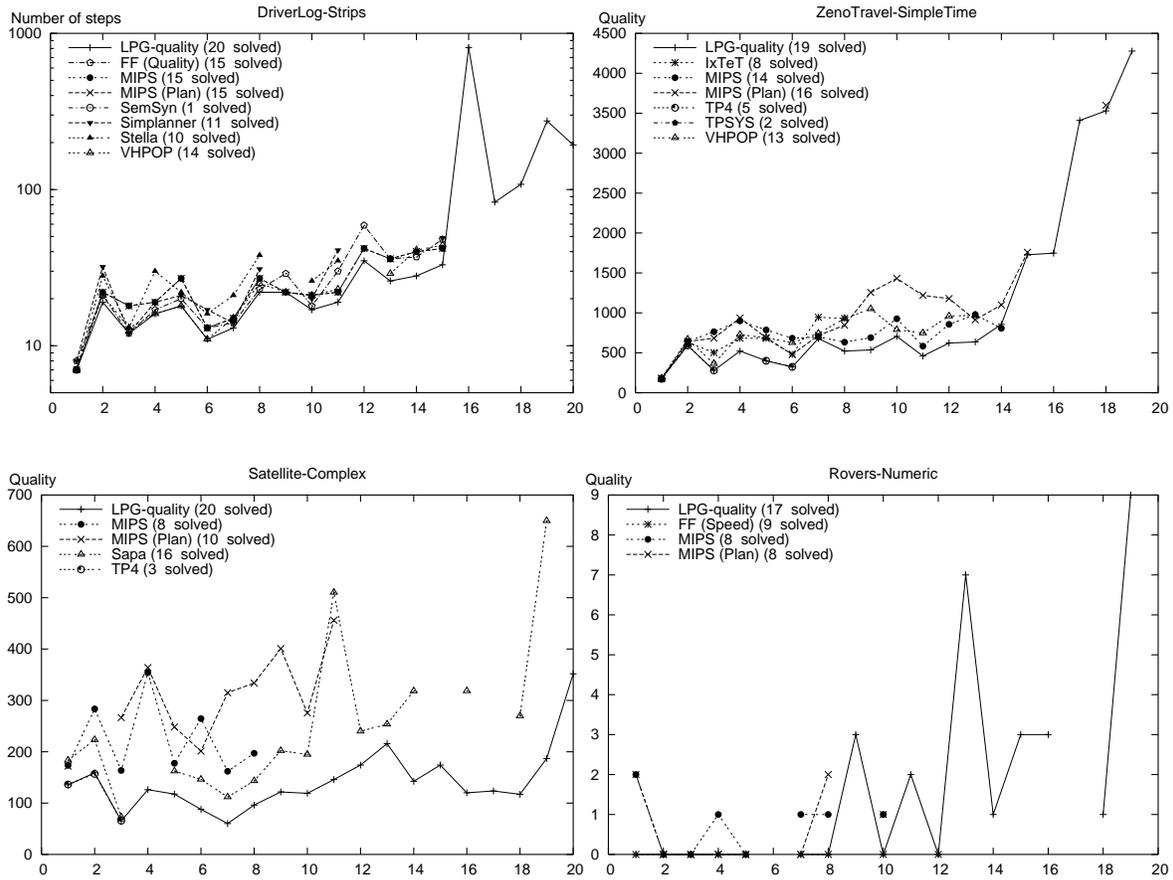

Figure 10: Quality of plans computed by the fully-automated planners of the 3rd IPC for the domains DriverLog Strips, ZenoTravel SimpleTime, Satellite Complex, Depots Time and Rovers Numeric. In order to improve readability, the plot for DriverLog-Strips is given in logarithmic scale.

The results in these and in the following plots are mostly self explanatory. In the temporal and complex variants LPG-speed is often one or more orders of magnitude faster than the SuperPlanner. In the Strips variant the SuperPlanner is faster for the smallest problems, but it is generally slower for the largest ones. In the Numeric variant the SuperPlanner is faster, but our planner produces solutions of better quality. Regarding LPG-quality, in all variants except Satellite Strips our planner performs much better than the SuperPlanner. In the Strips variant, the quality of the plans produced by LPG-quality is approximately the same as the quality of the plans generated by the SuperPlanner.

Concerning the quality of the solutions computed by LPG-speed and the CPU-time required by LPG-quality, as we have described in Section 3.6, it is important to note that LPG produces additional intermediate solutions, that for clarity are not shown in these plots. It is not surprising that very often plan quality for LPG-speed is poor with respect to plan quality for LPG-quality, and that the CPU-time required by LPG-quality is much higher that the CPU-time required by LPG-speed. Things become less clear if we compare plan quality for LPG-speed (or CPU-time for LPG-quality) and plan quality for the speed version of the SuperPlanner (or CPU-time for the quality version of the SuperPlanner).

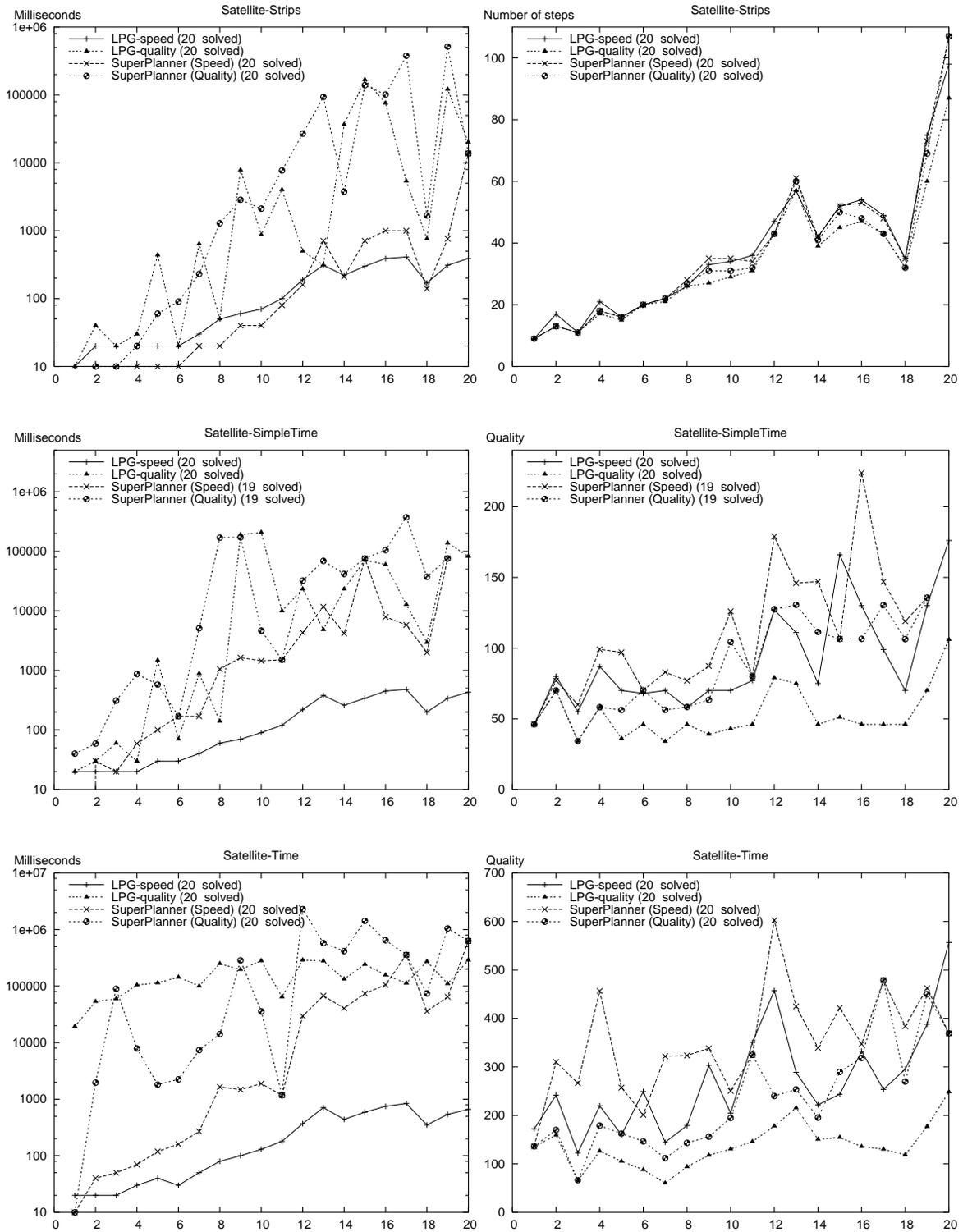

Figure 11: Performance of LPG-speed (left plots) and LPG-quality (right plots) compared with the SuperPlanner (speed and quality versions) in **Satellite** Strips, SimpleTime and Time.

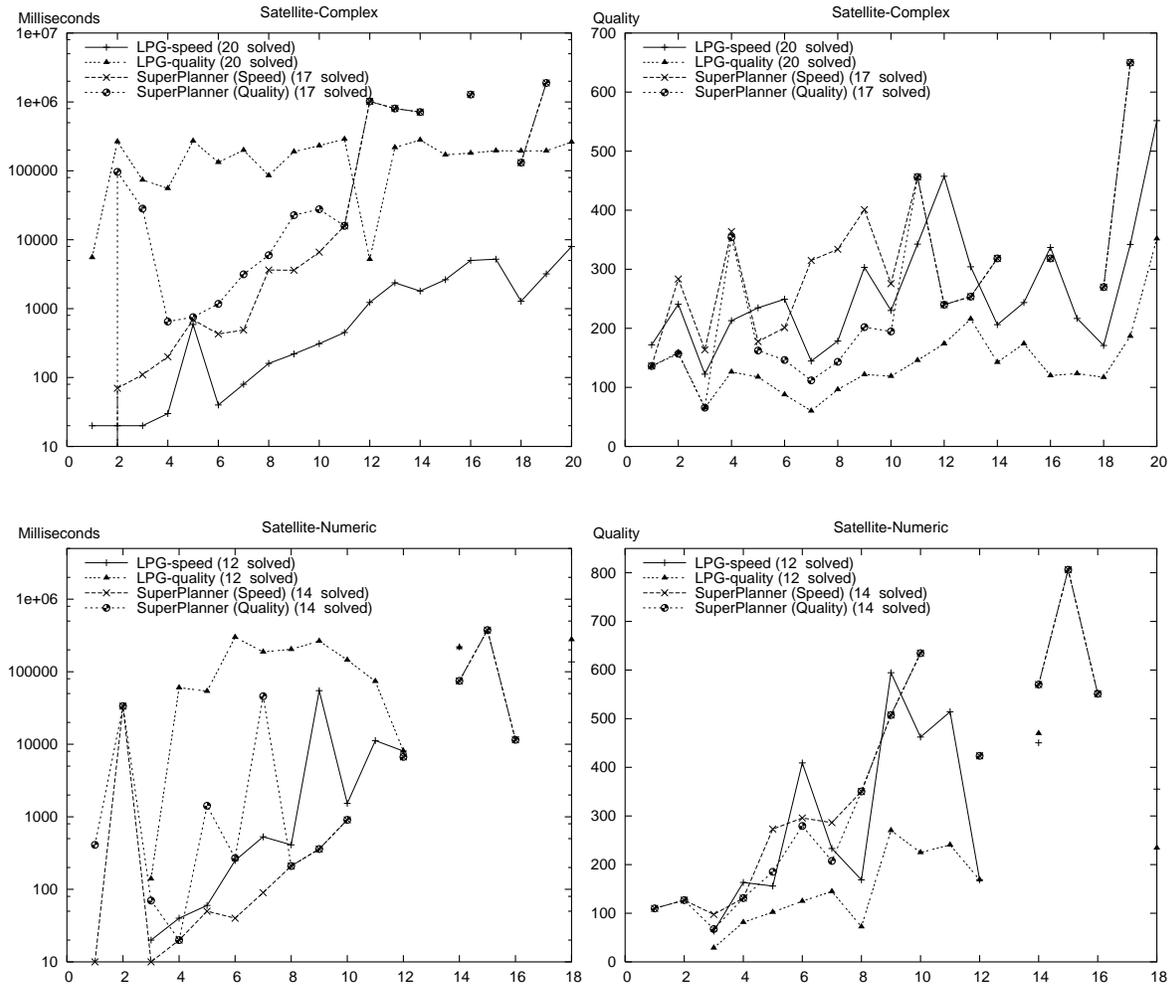

Figure 12: Performance of LPG-speed (left plots) and LPG-quality (right plots) compared with the SuperPlanner (speed and quality versions) in **Satellite** Complex and Numeric.

Given the any time nature of LPG, obviously there is a tradeoff between plan quality and speed. The more CPU-time the planner is allowed to run, the better the last solution generated. If we want to study this tradeoff experimentally, we need to consider not only the first and last solutions that LPG found in the competition tests, but also the interme-diate solutions. In fact, we observed that in several cases LPG generates an intermediate solution that has quality better than or similar to the quality of the best plan generated by the SuperPlanner, *and* that requires less or no more CPU-time than the SuperPlanner. In Figure 13 we give some support to our claim (a detailed analysis of all intermediate solutions generated by our planner is beyond the scope of this paper). The plots in this figure show CPU-time and quality of all plans generated by LPG (five runs) and by the SuperPlanner for some problems in the Satellite domain. In Satellite-SimpleTime-pfile15 and Satellite-Time-pfile6 (first two plots of Figure 13) LPG's first solutions (LPG-speed) require less CPU-time than the first solution found by the SuperPlanner. On the other hand, the quality of LPG-speed's solutions are worse than the quality of the solution found by the speed version of the SuperPlanner (see also the corresponding plots of Figure 11, keeping in mind that they are derived from median values over five runs). However, Figure 13 shows that for these two problems LPG-speed generates additional intermediate solutions that have quality better than the solutions found by the SuperPlanner, and that still require less CPU-time. Moreover, the third and fourth plots of Figure 13 show that in `Satellite-Strips-pfile9` and `Satellite-Numeric-pfile3` the SuperPlanner is faster than LPG-speed, but LPG finds intermediate solutions of quality better than the best solution of the SuperPlanner using less CPU-time than the SuperPlanner and LPG-quality.

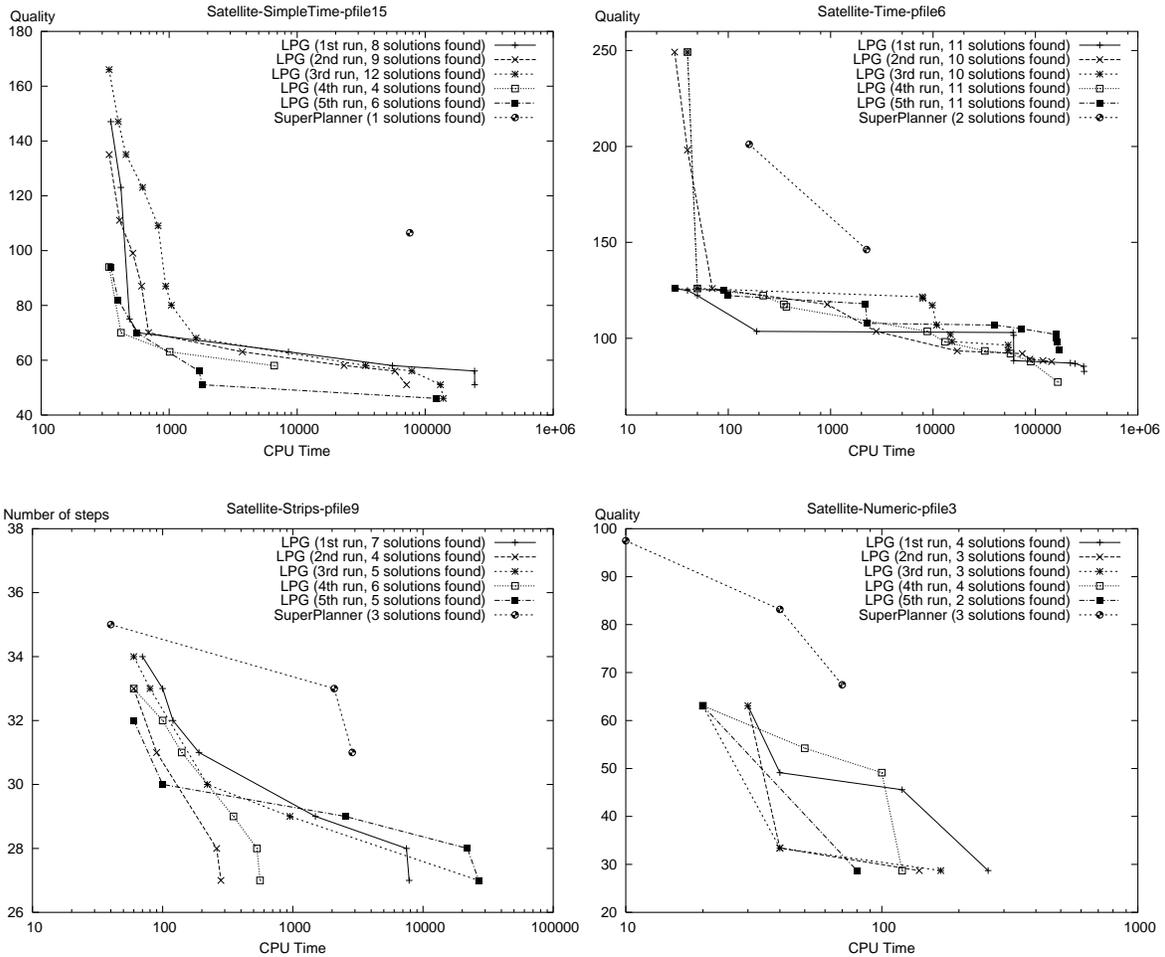

Figure 13: Plan quality and the corresponding CPU-milliseconds (logarithmic scale) for the solutions found by LPG (five runs) and the SuperPlanner for four problems in Satellite Strips, SimpleTime, Time and Numeric.

Since our main focus in this paper is temporal planning, it is interesting to compare LPG and the SuperPlanner in the Time variant of all competition domains. The detailed results of this comparison are given in Appendix B. As shown by the plots in this appendix, LPG-speed is usually faster than the SuperPlanner, and it always solves a larger number of problems, except in `ZenoTravel`, where our planner solves one problem less than the SuperPlanner. This problem was solved by MIPS, another planner of the 3rd IPC that performed well in the temporal domains (Edelkamp, 2002). The percentage of the problems solved by LPG-speed is 95.1%, while those solved by the SuperPlanner is 77.5%. The percentage of the problems in which our planner is faster is 81.4%, the percentage in which it is slower is 13.7%.

Regarding LPG-quality, generally in these domains the quality of the best plans produced by our planner is similar to the quality of the plans generated by the SuperPlanner, with some significant differences in `ZenoTravel`, where in a few problems the SuperPlanner performs better, and in `Satellite`, where our planner always performs better. Overall, in the Time variant of all the domains the percentages of the problems in which our planner produces a solution of better/worse quality are the same as the percentages of the problems in which LPG-speed is faster/slower.

We have also analyzed the performance of LPG with respect to the SuperPlanner for all other domains and problems attempted. Appendices C and D give summary results.[25] As for the Time-problems, in the SimpleTime problems LPG solves more problems than the SuperPlanner, and the percentages of problems in which LPG-speed and LPG-quality perform better than SuperPlanner are even higher than the corresponding percentages for the Time variants. In the Numeric and Strips problems, on average LPG-speed is less efficient than the SuperPlanner. This is mainly due to the generally good performance of FF in these domains. However, note that LPG-quality on average is better than the SuperPlanner in every domain except the Strips version of `ZenoTravel`.

Overall, considering all problems attempted, LPG-speed performs better/worse than the SuperPlanner in 55.8/38.1% of the problems, while LPG-quality performs better/worse in 71/11.6% of the problems.

Finally, we ran our planner on some of the large problems that were used to test the hand-coded planners in the 3rd IPC. In this experiment LPG was tested using a PC Pentium III, 500 MHz, with 1 Gbyte of RAM, which is more than two times slower than the machine used for testing the hand-coded planners. Of course, we did not expect to solve these problems more efficiently than the hand-coded planners. This experiment was aimed at testing how far we are from planners exploiting domain knowledge.

Figure 14 shows plots comparing the performance of LPG and the competing hand-coded planners for the two temporal variants of `Rovers`. LPG solved 38 of the 40 problems attempted. In terms of plan quality, very often LPG-quality generates plans that are nearly as good as those computed by the hand-coded planners, especially in `Rovers-SimpleTime-HandCoded`. Interestingly, given that the machine used to test LPG was slower, in this domain LPG-speed appears to perform slightly better than SHOP2 (Nau, Au, Ilghami, Kuter, Murdock, Wu, & Yaman, 2003). In `Rovers-Time-HandCoded` LPG-speed can solve most of the problems, but it does not perform as well. It remains an open question whether further research can reduce this gap significantly, but we are optimistic about this.

---

25. The paper in this issue by Long and Fox (2003) presents a detailed statistical analysis of all official results of the 3rd IPC.

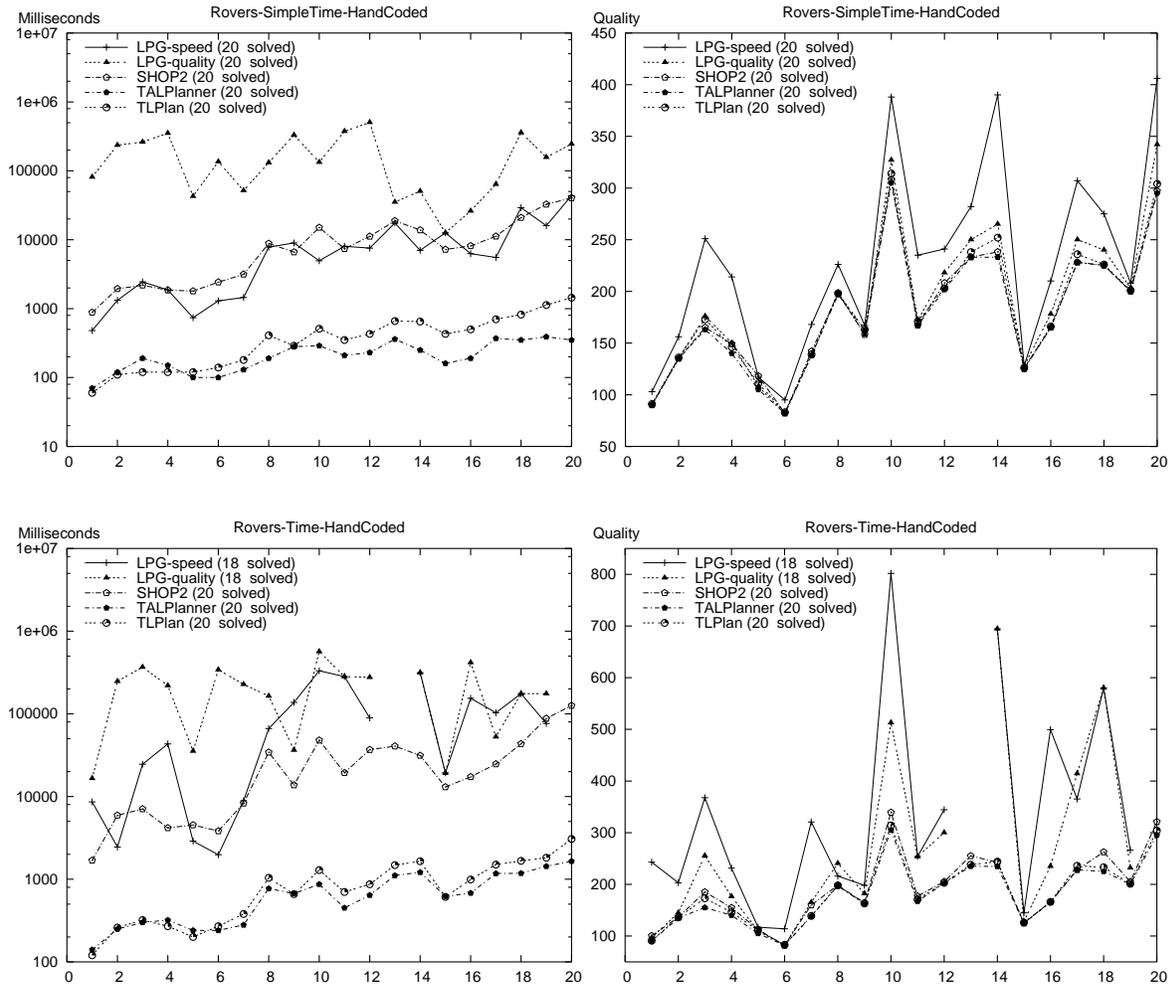

Figure 14: Performance of LPG in two temporal domains designed for hand-coded planners competing at the 3rd IPC. LPG was tested on a machine that is more than two times slower than the machine used to test the other planners.

## 5. Conclusions and Future Work

We have presented some new techniques for planning in PDDL2.1 domains that are implemented in LPG, an incremental (any time) planner producing multi-criteria quality plans. LPG was given an award for "distinguished performance of the first order" at the 3rd International Planning Competition, and additional experimental results presented in this paper give further evidence of the high performance of our system.

Other related techniques that are implemented in LPG, but not described here, concern: the restriction of the search neighborhood when it contains many elements, and their evaluation can slow down the search excessively; different strategies to choose the inconsistency to handle at each search step; the use of Lagrange multipliers in the action evaluation function (Gerevini & Serina, 2002, 2003).

We have already mentioned some directions that we are pursuing to improve our system. These include, in particular, an extension of our algorithm for computing reachability information taking account of numerical preconditions and goals (a recent related method has been proposed by Hoffmann (2003)). In addition, we intend to test other local search strategies for action graphs based on the use of a "tabu list" (Gerevini & Serina, 1999), and further types of graph modifications, some of which were implemented in the previous version of LPG (Gerevini & Serina, 2002). This might be especially important for improving the incremental plan-quality process. Another possible improvement of this process that is worth investigating is the use of dynamic coefficients to weigh the terms of the action evaluation function. When we start a new search for a plan of better quality, the weights of the terms representing the execution and temporal costs could be increased with respect to the term representing the search cost. This could guide the search towards plans better than those already derived, which is the purpose of the incremental process.

Finally, other directions for improving temporal planning in LPG concern the treatment of a richer temporal representation to handle upper and lower bounds on the possible action durations, as well as the integration of temporal reasoning techniques to deal with temporal constraints between actions similar to those that can be stated using Allen's Interval Algebra (Allen, 1983) or STP-constraints (Dechter, Meiri, & Pearl, 1991).

## Acknowledgments

The development of LPG has been carried out with the valuable support of several undergraduate students. Their help in the implementation and testing of LPG, before and during the competition in Toulouse, was very important. We would like to thank especially Marco Lazzaroni and Sergio Spinoni. Thanks also to Fabrizio Morbini, Valerio Lorini, and Stefano Orlandi for their support during the competition, and to all students of the AI class at the University of Brescia who made a contribution to this project. Finally, thanks to Maria Fox and Derek Long for their hard work in the organization of the 3rd IPC, to Jörg Hoffmann who made the source code of FF available (the parser and some data structures in LPG are based on extensions of Jörg's code), and to David Smith and the anonymous reviewers for their helpful comments.

## Appendix A: Mutex Relations in LPG and Related Work

LPG precomputes a set of mutex relations for the input planning problem using the two algorithms given in Figure 15, where $Add(a)$ denotes the set of the positive effects of $a$, $Del(a)$ the set of its negative effects, and $Pre(a)$ the set of its preconditions. ComputeMutexFacts derives a set of mutex relations between facts, that are used by ComputeMutexActions to compute a set of relations between actions. The correctness of this second algorithm is obvious since it just applies the original definition of mutex relation (Blum & Furst, 1997).

ComputeMutexFacts iteratively constructs a set $M$ of *potential* mutex relations and the set $F$ of all possible facts for the planning problem under consideration. At each iteration we consider every possible action $a$ (step 5) to possibly generate a set of new potential mutex relations (steps 7–11), and to possibly invalidate other potential mutex relations that have already been formulated (steps 12–18). The algorithm terminates when all possible facts have been considered ($F^* = F$), and no new potential mutex relations can be generated ($M^* = M$). When the algorithm terminates, $M$ contains a set of persistent mutex relations between facts. A mutex relation $m$ in $M$ is *persistent* if there is no state that can be reached from the initial state of the problem, using the operators of the domain under consideration, in which the facts of $m$ are both true. All mutex relations in the fixed-point level of a traditional planning graph are persistent.

Given an action $a$, two facts $f_1$ and $f_2$ form a potential mutex relation $m$ if (1) one of them is a positive effect of $a$ and the other is a negative effect (steps 7–9), or (2) one of them is a positive effect of $a$ and the other is (potentially) mutually exclusive with a precondition of $a$ (steps 7, 10 and 11). (1) is a natural way of hypothesizing mutex relations that is used also by Gerevini and Schubert (1998). (2) is based on the observation that, if $f_1$ is an effect of $a$, $p \in Pre(a)$, $f_2 \notin Add(a)$, and $f_2$ is mutually exclusive with $p$, then in any state resulting from the application of $a$ to a reachable state, $f_2$ and $f_1$ cannot be both true.

A potential mutex relation $m \in M$ between $f_1$ and $f_2$ becomes invalid if (1) there exists an action containing the two facts of $m$ among its positive effects (steps 13–14), or $f_1$ ($f_2$) is an add-effect of an action $a$, $f_2$ ($f_1$) is not deleted by $a$, and $f_2$ ($f_1$) is (potentially) mutually exclusive with no precondition of $a$ (steps 15–18). The first case if obvious, while the second can be explained as follows. If $f_1$ is a positive effect of $a$, and we cannot exclude that $f_2$ is true in a state where $a$ can be applied, then $f_2$ could persist from this state to the state produced by $a$ (similarly if $f_2$ is a positive effect of $a$).

Note that LPG handles negative preconditions as proposed by Koehler *et al.* (1997), i.e., no explicit atomic negation is available in LPG's language. Instead we model atomic negation by introducing an additional predicate $not\text{-}p(x)$ if $\neg p(x)$ is needed and by formulating add and delete effects correspondingly (this guarantees than $not\text{-}p(x)$ and $p(x)$ are mutex).

The next theorem states the correctness of our algorithms.

**Theorem** ComputeMutexFacts *and* ComputeMutexActions *correctly compute a set of persistent mutex relations between facts and actions respectively.*

**Proof.** Correctness of ComputeMutexActions is obvious, since it is a direct consequence of the definition of persistent mutex relation between actions. Correctness of ComputeMutexFacts follows from the two conditions under which a potential mutex relation is made

ComputeMutexFacts($I, \mathcal{O}$)

  *Input*: An initial state ($I$) and all ground operator instances ($\mathcal{O}$);
  *Output*: A set of persistent mutex relations between facts ($M$).

```
1   F* ← I; F ← ∅;
2.  M ← ∅; M* ← ∅; A ← ∅;
3.  while F* ≠ F ∨ M* ≠ M
4.      F ← F*; M ← M*;
5.      forall a ∈ O such that Pre(a) ⊆ F* and ¬(∃p, q ∈ Pre(a) ∧ (p, q) ∈ M*)
6.          New(a) ← Add(a) − F*;
7.          forall f ∈ New(a)
8.              forall h ∈ Del(a)
9.                  M* ← M* ∪ {(f, h), (h, f)};      /* Potential mutex relation */
10.             forall (p, q) ∈ M* such that p ∈ Pre(a) and q ∉ Del(a)
11.                 M* ← M* ∪ {(f, q), (q, f)};      /* Potential mutex relation */
12.         if a ∉ A then
13.             forall p, q ∈ Add(a) such that (p, q) ∈ M*
14.                 M* ← M* − {(p, q), (q, p)};      /* Invalid mutex relation */
15.         L ← Add(a) − New(a);
16.         forall (i, q) ∈ M* such that i ∈ L
17.             if q ∉ Del(a) ∧ ¬(∃p ∈ Pre(a) ∧ (p, q) ∈ M*) then
18.                 M* ← M* − {(i, q), (q, i)};      /* Invalid mutex relation */
19.         F* ← F* ∪ New(a);
20.         A ← A ∪ {a};
21. return M.
```

ComputeMutexActions($M, \mathcal{O}$)

  *Input*: A set of mutex relations between facts ($M$) and all ground operator instances ($\mathcal{O}$);
  *Output*: A set of persistent mutex relations between actions ($N$).

```
1.  N ← ∅; O* ← O extended with the no-op of every fact;
2.  forall (p, q) ∈ M
3.      forall a ∈ O* such that p ∈ Pre(a)
4.          forall b ∈ O* such that q ∈ Pre(b)
5.              N ← N ∪ {(a, b), (b, a)};      /* Competing needs */
6.  forall a ∈ O*
7.      forall p ∈ Pre(a)
8.          forall b ∈ O such that p ∈ Del(b)
9.              N ← N ∪ {(a, b), (b, a)};      /* Interference */
10.     forall p ∈ Add(a)
11.         forall b ∈ O such that p ∈ Del(b)
12.             N ← N ∪ {(a, b), (b, a)};      /* Inconsistent effects */
13. return N.
```

Figure 15: lpg's algorithms for computing the mutex relations.

invalid by the algorithm, and it can be proved by an inductive argument on the number $k$ of actions applied to reach a state $S$ from the initial state.

*Induction base $(k = 0)$.* It is easy to see that each element $m$ in the output set $M$ is a valid mutex relation for the initial state $(S = I)$, because the algorithm cannot formulate mutex relations involving two facts that are both true in the initial state.

*Induction hypothesis $(k = n)$.* Suppose that any element $m$ in the output set $M$ is a valid mutex relation in any state reached by the application of $n$ actions $(n \geq 1)$.

*Induction step $(k = n + 1)$.* Assume that there exists an element $m$ in the output set $M$ that is not a valid mutex relation in a state $S$ reachable by applying a sequence of $n + 1$ actions (because the two facts $f_1$ and $f_2$ of $m$ are both true in $S$), and let $a_{n+1}$ be the last action in this sequence. By the inductive assumption this can happen only if (i) $f_1$ and $f_2$ are both positive effects of $a_{n+1}$, or (ii) $f_1$ $(f_2)$ is an add-effect of $a_{n+1}$, $f_2$ $(f_1)$ is not deleted by $a_{n+1}$, and $f_2$ $(f_1)$ is true in the state $S'$ where $a_{n+1}$ is applied. Case (i) is ruled out by steps 13–14 of ComputeMutexActions. Regarding case (ii), since we are assuming that $S'$ is a reachable (consistent) state where $f_2$ $(f_1)$ is true and $a_{n+1}$ can be applied, there must exist no precondition $p$ of $a_{n+1}$ that is mutex with $f_2$ $(f_1)$. Moreover, by the inductive assumption $(p, f_2)$ $((p, f_1))$ cannot belong to the output $M$-set – if some iteration of the algorithm adds the potential mutex relation between $p$ and $f_2$ $(f_1)$ to $M$, then it must be the case that it is then removed from $M$. It follows that, if some iteration adds $(f_1, f_2)$ to $M$, steps 16–18 will then remove it from $M$, contrary to our assumption that $m$ belongs to the output $M$-set.

Termination of the two algorithms is guaranteed because there is always a finite maximum number of different facts, actions and potential mutex relations. □

Smith and Weld proposed the notion of "eternal mutex" (*emutex*) as a mutex relation that persists for all time (Smith & Weld, 1999). According to their definition of emutex, our persistent mutex relations between facts and between actions subsume theirs. Concerning emutex relations between an action and a fact, Smith and Weld consider an action $a$ with effect $p$ emutex with a fact $p$, while we do not consider $a$ and *no-op(p)* persistently mutex.

Bonet and Geffner (2001) proposed a method for deriving a set of mutex relations between facts that has some similarities with ours. Both methods are based on hypothesizing a set of pairs of mutex facts that are then possibly eliminated from the set according to certain conditions on the preconditions and effects of the actions. However, there are also some significant differences. While Bonet and Geffner compute an initial large set $M_0$ of candidate mutex pairs, and then prune it, ComputeMutexFacts incrementally constructs and verifies the set $M$ through a forward process. The conditions under which a pair of facts is in $M_0$ are different from the conditions used by ComputeMutexFacts to create $M$ (especially the condition in step 10). Moreover, our algorithm generates and tests the pairs of $M$ considering only applicable actions (i.e., actions with all preconditions in $F^*$ and with non-mutex preconditions), while Bonet and Geffner derive $M_0$ using every operator instance. Finally, their paper does not contain algorithmic details about the identification of "bad pairs" in $M_0$, and there is no formal proof of correctness.

For problems involving a very high number of actions, precomputing mutex relations could be computationally very expensive. In order to cope with these cases, the user of LPG can set an option of the planner (lowmemory) for computing the mutex relations between actions at search time (while those between facts and between actions and no-ops are still precomputed). Preprocessing with `lowmemory` on becomes faster and requires much less memory, but each search step becomes slower. For this reason in the current version of LPG this option is recommended only when the precomputation of mutex relations between actions is prohibitive. This was never the case for the test problems of the 3rd IPC designed for the fully-automated planners, but for some of the problems designed for the hand-coded planners, like those of the domain `Satellite` Hand-Coded, the use of this option is necessary. Currently we are studying an alternative method for computing (persistent) mutex relations during search based on the use of state invariants automatically derived by existing domain analysis tools, such as DISCOPLAN (Gerevini & Schubert, 1998) or TIM (Fox & Long, 1998a). A similar method has been proposed by Fox and Long (2000).

Finally, for domains involving numerical preconditions and effects, the set of mutex relations between actions computed by the algorithms of Figure 15 is extended using the definition of mutex relations for numeric domains given by Fox and Long (2003).

# Appendix B: LPG and the SuperPlanner in the Time variant of the competition domains

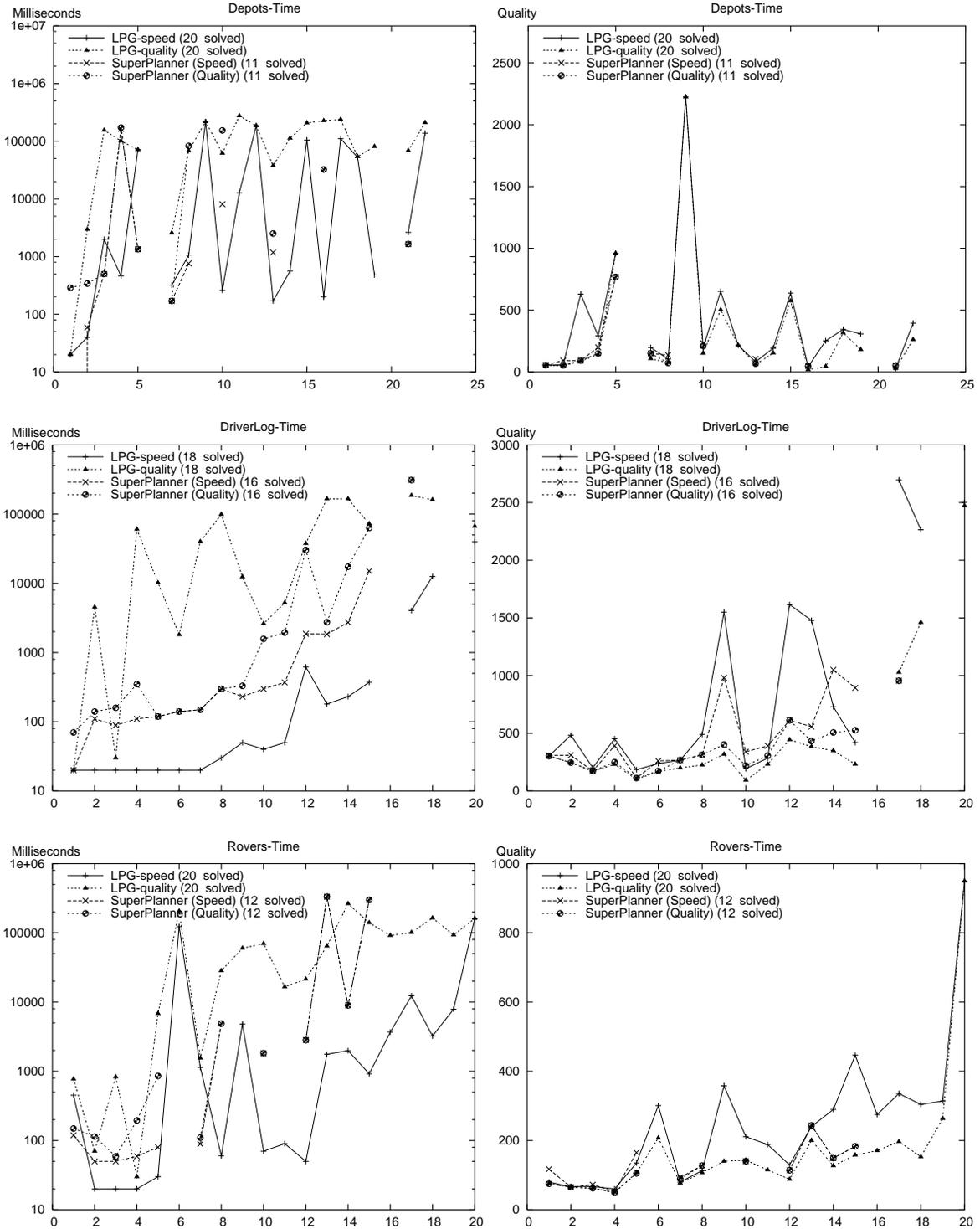

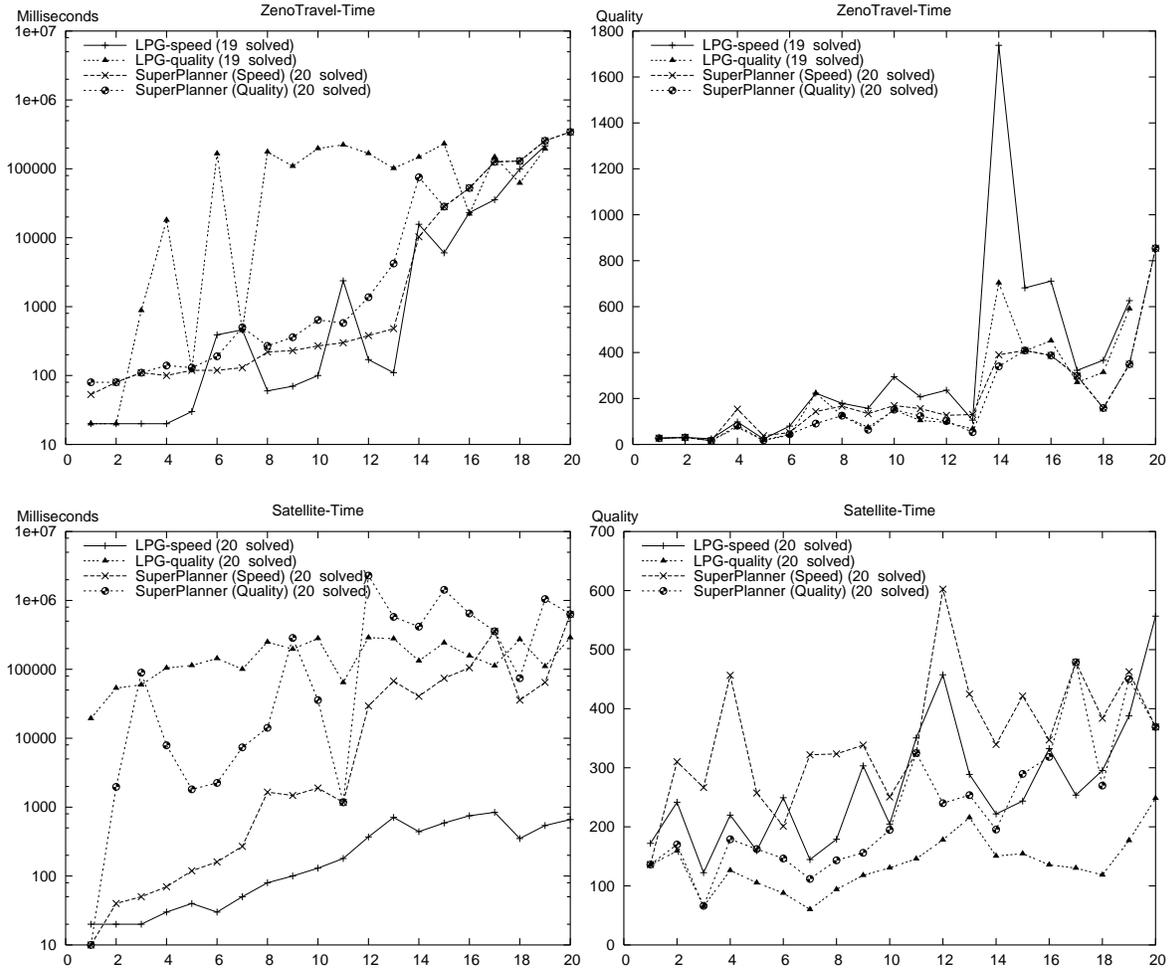

## Appendix C: Comparison of LPG-speed and the SuperPlanner

The following table shows the performance of LPG-speed and the SuperPlanner in every variant of every domain tested using our local search techniques. The two systems are compared in terms of: number of problems solved (2nd and 3rd columns); number of problems in which LPG-speed is faster/slower than the SuperPlanner (4th/6th columns); number of problems in which LPG-speed is much faster/slower than the SuperPlanner (5th/7th columns). A system was considered much faster than the other one when the CPU-time required by the first was at least one order of magnitude lower than the second. When a planner was not able to find a solution, the required CPU-time was considered infinite.

| Domain | Problems solved by LPG | Problems solved by the Super-Planner | LPG better then the Super-Planner | LPG much better then the Super-Planner | LPG worse than the Super-Planner | LPG much worse than the Super-Planner |
|---|---|---|---|---|---|---|
| **Strips** | | | | | | |
| Depots | 22 (100%) | 22 (100%) | 6 (27.3%) | 0 (0%) | 16 (72.7%) | 5 (22.7%) |
| DriverLog | 20 (100%) | 15 (75%) | 7 (35%) | 5 (25%) | 12 (60%) | 1 (5%) |
| Rovers | 20 (100%) | 20 (100%) | 4 (20%) | 0 (0%) | 14 (70%) | 3 (15%) |
| Satellite | 20 (100%) | 20 (100%) | 6 (30%) | 1 (5%) | 14 (70%) | 2 (10%) |
| ZenoTravel | 19 (95%) | 20 (100%) | 0 (0%) | 0 (0%) | 20 (100%) | 12 (60%) |
| Total | 99% | 95.1% | 22.5% | 5.9% | 74.5% | 22.5% |
| **Simple-time** | | | | | | |
| Depots | 21 (95.5%) | 11 (50%) | 18 (81.8%) | 14 (63.6%) | 3 (13.6%) | 0 (0%) |
| DriverLog | 18 (90%) | 16 (80%) | 15 (75%) | 6 (30%) | 3 (15%) | 2 (10%) |
| Rovers | 20 (100%) | 10 (50%) | 17 (85%) | 12 (60%) | 1 (5%) | 1 (5%) |
| Satellite | 20 (100%) | 19 (95%) | 18 (90%) | 12 (60%) | 1 (5%) | 1 (5%) |
| ZenoTravel | 19 (95%) | 16 (80%) | 18 (90%) | 9 (45%) | 1 (5%) | 0 (0%) |
| Total | 96% | 70.6% | 83.4% | 51.9% | 8.8% | 3.9% |
| **Time** | | | | | | |
| Depots | 20 (90.9%) | 11 (50%) | 14 (63.6%) | 12 (54.5%) | 6 (27.3%) | 2 (9.1%) |
| DriverLog | 18 (90%) | 16 (80%) | 17 (85%) | 6 (30%) | 0 (0%) | 0 (0%) |
| Rovers | 20 (100%) | 12 (60%) | 18 (90%) | 13 (65%) | 2 (10%) | 1 (5%) |
| Satellite | 20 (100%) | 20 (100%) | 19 (95%) | 12 (60%) | 1 (5%) | 0 (0%) |
| ZenoTravel | 19 (95%) | 20 (100%) | 15 (75%) | 0 (0%) | 5 (25%) | 1 (5%) |
| Total | 95.1% | 77.5% | 81.4% | 42.1% | 13.7% | 3.9% |
| **Numeric** | | | | | | |
| Depots | 21 (95.5%) | 20 (90.9%) | 8 (36.4%) | 2 (9.1%) | 12 (54.5%) | 2 (9.1%) |
| DriverLog | 18 (90%) | 16 (80%) | 7 (35%) | 3 (15%) | 10 (50%) | 3 (15%) |
| Rovers | 17 (85%) | 9 (45%) | 10 (50%) | 8 (40%) | 7 (35%) | 3 (15%) |
| Satellite | 12 (60%) | 14 (70%) | 2 (10%) | 2 (10%) | 14 (70%) | 5 (25%) |
| ZenoTravel | 20 (100%) | 20 (100%) | 0 (0%) | 0 (0%) | 20 (100%) | 5 (25%) |
| Total | 83.6% | 77.4% | 26.4% | 14.7% | 61.8% | 17.6% |
| **Complex** | | | | | | |
| Satellite | 20 (100%) | 17 (85%) | 19 (95%) | 14 (70%) | 1 (5%) | 1 (5%) |
| **Hard-numeric** | | | | | | |
| DriverLog | 20 (100%) | 16 (80%) | 12 (60%) | 5 (25%) | 8 (40%) | 2 (10%) |
| Total | 94.6% | 80.3% | 55.8% | 30.3% | 38.1% | 11.6% |

## Appendix D: Comparison of LPG-quality and the SuperPlanner

The following table shows the performance of LPG-quality and the SuperPlanner in every variant of every domain tested using our local search techniques. The two systems are compared in terms of: number of problems solved (2nd and 3rd columns); number of problems in which the quality of the solution computed by LPG is better/worse than the solution computed by the SuperPlanner (4th/6th columns); number of problems in which the solution of LPG-quality is much better/worse than the solution of the SuperPlanner (5th/7th columns). A solution $\pi$ derived by a system is considered much better than the solution $\pi'$ for the same problem derived by the other system if the quality of $\pi$ is at least twice as good as the quality of $\pi'$, or if $\pi$ exists and $\pi'$ does not exist (because the system could not solve the corresponding problem). The quality of a plan is measured using the plan metric indicated in the problem specification, except for the Strips problems, where plan quality is defined as the number of actions. In all problems considered, the lower the value of the metric expression, the better the plan is.

| Domain | Problems solved by LPG | Problems solved by the Super-Planner | LPG better than the Super-Planner | LPG much better than the Super-Planner | LPG worse than the Super-Planner | LPG much worse than the Super-Planner |
|---|---|---|---|---|---|---|
| **STRIPS** | | | | | | |
| Depots | 22 (100%) | 22 (100%) | 12 (54.5%) | 0 (0%) | 4 (18.2%) | 0 (0%) |
| DriverLog | 20 (100%) | 15 (75%) | 14 (70%) | 5 (25%) | 0 (0%) | 0 (0%) |
| Rovers | 20 (100%) | 20 (100%) | 9 (45%) | 0 (0%) | 1 (5%) | 0 (0%) |
| Satellite | 20 (100%) | 20 (100%) | 13 (65%) | 0 (0%) | 0 (0%) | 0 (0%) |
| ZenoTravel | 19 (95%) | 20 (100%) | 3 (15%) | 0 (0%) | 9 (45%) | 1 (5%) |
| Total | 99% | 95.1% | 50% | 4.9% | 13.7% | 0.9% |
| **Simple-time** | | | | | | |
| Depots | 21 (95.5%) | 11 (50%) | 19 (86.4%) | 11 (50%) | 1 (4.5%) | 0 (0%) |
| DriverLog | 18 (90%) | 16 (80%) | 17 (85%) | 3 (15%) | 1 (5%) | 0 (0%) |
| Rovers | 20 (100%) | 10 (50%) | 20 (100%) | 10 (50%) | 0 (0%) | 0 (0%) |
| Satellite | 20 (100%) | 19 (95%) | 20 (100%) | 7 (35%) | 0 (0%) | 0 (0%) |
| ZenoTravel | 19 (95%) | 16 (80%) | 17 (85%) | 3 (15%) | 2 (10%) | 0 (0%) |
| Total | 96% | 70.6% | 91.2% | 33.3% | 3.9% | 0% |
| **Time** | | | | | | |
| Depots | 20 (90.9%) | 11 (50%) | 17 (77.3%) | 9 (40.9%) | 2 (9.1%) | 0 (0%) |
| DriverLog | 18 (90%) | 16 (80%) | 17 (85%) | 4 (20%) | 1 (5%) | 0 (0%) |
| Rovers | 20 (100%) | 12 (60%) | 18 (90%) | 8 (40%) | 2 (10%) | 0 (0%) |
| Satellite | 20 (100%) | 20 (100%) | 20 (100%) | 5 (25%) | 0 (0%) | 0 (0%) |
| ZenoTravel | 19 (95%) | 20 (100%) | 11 (55%) | 0 (0%) | 9 (45%) | 3 (15%) |
| Total | 95.1% | 77.4% | 81.4% | 25.5% | 13.7% | 2.9% |
| **Numeric** | | | | | | |
| Depots | 21 (95.5%) | 20 (90.9%) | 10 (45.5%) | 1 (4.5%) | 8 (36.4%) | 1 (4.5%) |
| DriverLog | 18 (90%) | 16 (80%) | 15 (75%) | 2 (10%) | 0 (0%) | 0 (0%) |
| Rovers | 17 (85%) | 9 (45%) | 8 (40%) | 8 (40%) | 0 (0%) | 0 (0%) |
| Satellite | 12 (60%) | 14 (70%) | 12 (60%) | 7 (35%) | 4 (20%) | 4 (20%) |
| ZenoTravel | 20 (100%) | 20 (100%) | 9 (45%) | 1 (5%) | 6 (30%) | 3 (15%) |
| Total | 86.3% | 77.4% | 52.9% | 18.6% | 17.6% | 7.8% |
| **Complex** | | | | | | |
| Satellite | 20 (100%) | 17 (85%) | 19 (95%) | 9 (45%) | 1 (5%) | 0 (0%) |
| **Hard-numeric** | | | | | | |
| DriverLog | 20 (100%) | 16 (80%) | 18 (90%) | 5 (25%) | 1 (5%) | 0 (0%) |
| Total | 94.6% | 80.3% | 71% | 21.9% | 11.6% | 2.7% |